\documentclass[fleqn,10pt,pdftex,cmex10,table]{wlscirep}
\usepackage[utf8]{inputenc}
\usepackage[T1]{fontenc}
\usepackage{xcolor,soul,framed} 
\colorlet{shadecolor}{yellow}
\DeclareGraphicsExtensions{.pdf,.jpeg,.png}

\usepackage{array}
\usepackage{mdwmath}
\usepackage{eqparbox}
\usepackage{url}
\usepackage{diagbox}
\usepackage{tabularx}
\usepackage{adjustbox}
\usepackage{rotating}
\usepackage{multirow}
\usepackage{subcaption}  
\usepackage{algorithm}
\usepackage{algpseudocode}
\usepackage{bm} 
\usepackage{amsmath} 
\usepackage{booktabs} 
\usepackage{graphicx}
\usepackage{subcaption}

\usepackage{amssymb}

\usepackage{tikz}

\usepackage{xcolor}
\usepackage{amssymb}
\usepackage{lscape}

    \title{Multilevel Determinants of Overweight and Obesity Among U.S. Children Aged 10–17: Comparative Evaluation of Statistical and Machine Learning Approaches Using the 2021 National Survey of Children’s Health}
    
\iftrue

\author[1,*]{Joyanta Jyoti Mondal}

\affil[1]{Department of Computer and Information Sciences, University of Delaware, United States}

\affil[1]{joyanta@udel.edu}

\affil[*]{Corresponding Author}

\fi

\begin{abstract}

\textbf{Background:}
Childhood and adolescent overweight and obesity remain major public health concerns in the United States and are shaped by behavioral, household, and community factors. Their joint predictive structure at the population level remains incompletely characterized.
\textbf{Objectives:}
The study aims to identify multilevel predictors of overweight and obesity among U.S. adolescents and compare the predictive performance, calibration, and subgroup equity of statistical, machine-learning, and deep-learning models.
\textbf{Data and Methods:}
We analyze 18,792 children aged 10–17 years from the 2021 National Survey of Children’s Health. Overweight/obesity is defined using BMI categories. Predictors included diet, physical activity, sleep, parental stress, socioeconomic conditions, adverse experiences, and neighborhood characteristics. Models include logistic regression, random forest, gradient boosting, XGBoost, LightGBM, multilayer perceptron, and TabNet. Performance is evaluated using AUC, accuracy, precision, recall, F1 score, and Brier score.
\textbf{Results:}
Discrimination range from 0.66 to 0.79. Logistic regression, gradient boosting, and MLP show the most stable balance of discrimination and calibration. Boosting and deep learning modestly improve recall and F1 score. No model was uniformly superior. Performance disparities across race and poverty groups persist across algorithms.
\textbf{Conclusion:}
Increased model complexity yields limited gains over logistic regression. Predictors consistently span behavioral, household, and neighborhood domains. Persistent subgroup disparities indicate the need for improved data quality and equity-focused surveillance rather than greater algorithmic complexity.
\end{abstract}


\begin{document}
\flushbottom
\maketitle
%
%
\thispagestyle{empty}

\section{Introduction}
\label{sec:intro}



Childhood overweight and obesity represent persistent public-health challenges in the United States and globally, contributing to elevated risks of metabolic disorders, functional impairments, and long-term cardiovascular disease. Obesity is a multifactorial condition characterized by abnormal or excessive fat accumulation resulting from an imbalance between energy intake and expenditure, often driven by high-calorie consumption and insufficient physical activity \cite{Kivrak2021-dl,Alvarez2011-ry}. According to the World Health Organization, in 2016 approximately 39\% of adults worldwide were overweight and 13\% were obese, with prevalence nearly tripling since 1975 \cite{obesity_overweight}. The anticipated monetary burden of obesity is substantial, and in the United States the direct costs of inactivity and obesity have been estimated to account for about 9.4\% of national healthcare expenditures \cite{Colditz1999-aw}. These trends impose substantial economic consequences, with obesity-related healthcare costs projected to account for 5–14\% of total health expenditures between 2020 and 2050 \cite{Oecd2019-ye}. Consistent with these trends, overweight and obesity remain highly prevalent across the U.S. population. Recent national estimates indicate that roughly one in five U.S. children and adolescents has obesity, with prevalence increasing with age, approximately 12.7\% among children aged 2–5 years, 20.7\% among those aged 6–11 years, and 22.2\% among adolescents aged 12–19 years \cite{CDCChildObesity}. Among adults, obesity prevalence remains similarly high, with an estimated 40.3\% of adults aged 20 years and older classified as obese between August 2021 and August 2023, and severe obesity affecting approximately 9.4\% of adults during the same period \cite{CDCAdultObesity}. In children and adolescents, individual behaviors such as diet quality, physical activity, sleep patterns, and screen exposure play central roles in shaping weight outcomes; however, risk is also structured by household stressors, socioeconomic conditions, family functioning, and neighborhood environments. An expanding literature highlights this multilevel architecture of obesity risk across behavioral, familial, and community domains, yet nationally representative analyses that evaluate these determinants simultaneously remain limited.

Research indicates that obesity risk extends beyond individual behaviors to encompass demographic characteristics, socioeconomic conditions, community infrastructure, and broader environmental contexts \cite{Gulu2022-lu,Reidpath2002-zu,Cohen2006-db,Lamerz2005-bq}. In lower socioeconomic populations, obesity prevalence has risen sharply; often by threefold or more; driven by urbanization, shifts in dietary patterns and food availability, and declining physical activity levels \cite{Ng2014-rs,Ford2008-ji}. Obesity is associated with increased mortality from non-communicable diseases such as type 2 diabetes, cardiovascular disease, and certain cancers, and has been linked to life expectancy reductions of up to 20 years \cite{Fontaine2003-ol,Berrington_de_Gonzalez2010-uf,Prospective_Studies_Collaboration2009-te,Pischon2008-ol}. Although largely preventable \cite{Jaacks2019-ur}, excess weight; particularly when it begins in childhood or adolescence; substantially elevates the risk of later-life conditions including cardiovascular disease \cite{Lavie2014-ah}, diabetes \cite{Koh-Banerjee2004-zf}, and asthma \cite{Ford2005-ml}. Accordingly, obesity is widely recognized as a multifactorial condition \cite{Bakhshi2015-nu}, shaped by socioeconomic status \cite{Sarlio-Lahteenkorva2004-bi}, occupation \cite{Bonauto2014-nb}, behavioral factors such as smoking \cite{John2005-mo}, and levels of physical activity \cite{Besson2009-wk}. At its core, overweight and obesity result from a sustained imbalance between energy intake and expenditure \cite{Popkin2012-br}, highlighting the central role of diet quality and physical activity as key modifiable determinants in obesity prevention and management \cite{Kahan2016-qd,Van_Baal2008-oq,Spiegelman2001-ij}.

Much of the existing literature in obesity research relies predominantly on traditional regression approaches or narrowly focused cohort studies, which are well suited for hypothesis testing but limited in their ability to capture nonlinear associations and contextual variation across diverse populations. In response, machine-learning methods have increasingly been applied to obesity prediction tasks, including studies using national survey data and pediatric cohorts, often demonstrating improved predictive performance relative to classical models \cite{Colmenarejo2020-ml, Ganie2025-hy}. However, many of these applications emphasize predictive accuracy rather than model interpretability, limiting their utility for understanding behavioral and socioeconomic mechanisms underlying obesity risk. Although interpretable machine-learning approaches such as SHapley Additive exPlanations (SHAP) have begun to appear in population health research, their use in obesity studies remains relatively limited \cite{Sun2024-ml, An2022-ai}. Deep-learning methods, while widely adopted in other areas of computational health research, remain underutilized in population-level obesity studies, in part because large health surveys involve heterogeneous, tabular data structures that are challenging to model while preserving transparency and interpretability \cite{Gupta2019-dl, An2022-ai}.


This study applies a comparative modeling strategy to examine overweight and obesity among U.S. children aged 10–17 years using the NSCH’s nationally representative dataset. By placing classical statistical models, tree-based machine-learning algorithms, and deep-learning architectures within a unified analytic framework, the study evaluates how these approaches capture nonlinear relationships, identify influential predictors, and differ in performance across demographic subgroups. The analysis incorporates interpretable attention-based modeling to clarify predictor importance and includes subgroup fairness evaluation to assess systematic performance disparities across sex, race, income level, and metropolitan status.

\begin{figure}[!ht]
    \centering
    \includegraphics[width=0.85\textwidth]{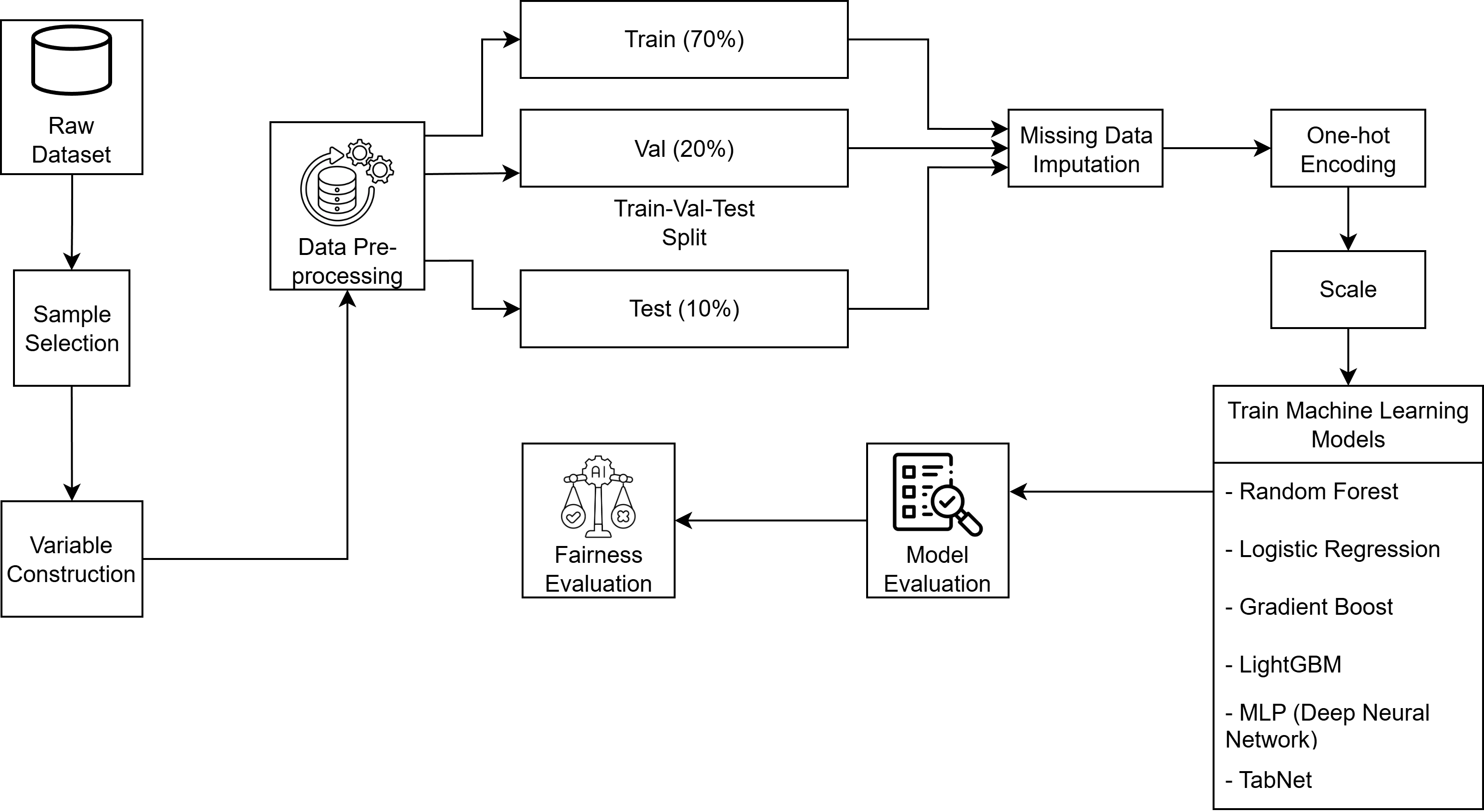}
    \caption{Overview of the work.
    }
    \label{fig:overview}
\end{figure}

Aligned with these aims, the study addresses the following research questions:

\begin{itemize}
\item RQ1: Which behavioral, household, and community-level factors are most strongly associated with overweight and obesity among U.S. children aged 10–17 years?
\item RQ2: How do classical statistical models, machine-learning algorithms, and deep-learning architectures differ in predictive performance when applied to a nationally representative pediatric dataset?
\item RQ3: Which predictors consistently emerge as influential across modeling approaches, and do nonlinear models reveal patterns not captured by logistic regression?
\item RQ4: Does predictive performance vary across demographic and socioeconomic subgroups, and what do these variations imply for fairness in childhood obesity prediction?
\end{itemize}

The contributions of this study are summarized as follows:

\begin{itemize}
\item Development of a comparative modeling framework that evaluates classical regression, tree-based machine-learning models, and deep-learning architectures to characterize predictors of childhood overweight and obesity in a large national dataset.

\item Integration of interpretable attention-based modeling to provide transparent estimates of behavioral, household, and community-level predictor importance, supporting evidence-based public-health interpretation.

\item Incorporation of subgroup fairness evaluation across sex, race, income, and metropolitan status to identify structural performance disparities and contextualize model behavior within broader social determinants.
\end{itemize}

The remainder of the paper is organized as follows: 
Section~\ref{Methodology} describes the models. Section \ref{experiment} describes the dataset, variable construction, and experimental design, including preprocessing.  Section~\ref{sec:results} presents descriptive analyses, performance comparisons, and interpretability results. Section~\ref{sec:fairness_evaluation} reports findings from the demographic fairness evaluation. Section~\ref{discussion} interprets the implications of these results. Section~\ref{Limitation} outlines methodological constraints. Finally, Section~\ref{Conclusion} summarizes the study’s contributions and identifies directions for future work.

\section{Methods}
\label{Methodology}




In this study, we evaluate a diverse set of predictive modeling approaches, including classical statistical methods, tree-based machine learning algorithms, gradient boosting frameworks, and deep neural network architectures. All models aim to classify the binary outcome $Y \in \{0,1\}$, where $Y=1$ indicates overweight/obesity. A high-level overview of our approach is shown in Figure~\ref{fig:overview}.

\subsection{Logistic Regression}

Logistic Regression is a baseline approach for modeling binary outcomes and serves as a natural starting point for epidemiological analyses. Its purpose is to estimate how different predictors relate to the probability that an event occurs, in this case, whether a child is overweight or obese.

The model assumes that each predictor contributes additively to the log-odds of the outcome. This allows each coefficient to be interpreted directly: it expresses how a one-unit change in a predictor increases or decreases the odds of the outcome while holding all other variables constant. This level of transparency is one of the main reasons logistic regression remains widely used in public-health research.

Even though the model is linear in its structure, it uses a nonlinear transformation, the logistic (sigmoid) function, to convert its predictions into valid probabilities between 0 and 1. The parameters are estimated by choosing the values that make the observed data most likely under the model.

A compact mathematical expression of the model is:


\[
\Pr(Y_i = 1 \mid x_i)
= \frac{1}{1 + e^{-(\beta_0 + \beta_1 x_i)}}
\]

Because the relationships between predictors and outcomes remain fully interpretable, logistic regression offers a clear benchmark against which more flexible machine-learning models can be compared.

\subsection{Random Forest}




Random Forest is an ensemble learning method built by combining many individual decision trees. Each tree is trained on a slightly different version of the dataset, created by sampling both observations and predictors. This intentional randomness prevents any single tree from dominating the model and reduces the risk of overfitting.

A decision tree works by repeatedly splitting the data into increasingly homogeneous groups based on predictor values. Although a single tree can capture complex nonlinear patterns and interactions, it tends to be unstable: small changes in the data can lead to very different trees. Random Forest addresses this instability by averaging the predictions across a large collection of trees. The ensemble behaves more smoothly than any of its components and typically generalizes better.



\subsection{Gradient Boosting}




Gradient Boosting is an iterative ensemble method that builds a predictive model by combining many weak learners; typically shallow decision trees, into a single, stronger model. Instead of training all trees independently, as in Random Forest, Gradient Boosting constructs trees sequentially. Each new tree focuses on the mistakes made by the earlier ones.

The model begins with a simple prediction, often the average outcome. It then examines the residuals; the parts of the data the model fails to explain, and fits a new tree that tries to correct those errors. This process repeats, with each tree learning to refine the predictions of the ensemble. The result is a model that can capture subtle nonlinearities and high–order interactions even when individual trees are weak learners.

As Gradient Boosting improves the model step by step, it is sensitive to noise if left unchecked. Learning–rate parameters and shallow tree depths are therefore used to keep the model from overfitting and to encourage gradual improvement. These controls make the algorithm stable while still highly flexible.


\subsection{XGBoost}




XGBoost (Extreme Gradient Boosting) is a modern, optimized implementation of gradient boosting designed specifically for speed, scalability, and strong performance on structured tabular data. While it follows the same general idea as traditional Gradient Boosting; building trees sequentially to correct previous errors, it introduces several engineering and algorithmic improvements that make it particularly effective in large and noisy datasets.

The key innovation is its use of a regularized objective function. XGBoost penalizes overly complex trees, which reduces overfitting and encourages models that generalize better to new data. It also uses second–order information (both gradients and curvature) when updating the model. This allows the algorithm to take more informed steps during training and improves stability, especially when predictors have different scales or distributions.

Another advantage is its handling of missing data. XGBoost learns default branching directions for missing values directly from the training process, allowing it to use available cases efficiently without requiring additional imputation strategies for tree-based components. This characteristic is useful for survey data where some variables naturally contain gaps.

On the computational side, XGBoost employs system-level optimizations such as parallel tree construction, cache-aware structures, and sparsity-aware algorithms. These features make it faster than classical boosting implementations, even when tuning many parameters or training on large samples.

\subsection{LightGBM}




LightGBM is another gradient–boosting framework designed for efficiency on large, high–dimensional tabular datasets. Like other boosting methods, it builds decision trees sequentially so that each new tree improves on the errors of the previous ones. What distinguishes LightGBM is the way it grows trees and handles data, allowing it to train faster while maintaining strong predictive performance.

Rather than splitting all possible features at each level, LightGBM uses a histogram–based algorithm. Continuous values are grouped into discrete bins before training begins. This reduces memory usage and speeds up split finding, especially when the dataset has many numeric variables. The model evaluates splits based on these bins rather than every unique value, which introduces minimal loss of precision but significantly reduces computation.

LightGBM also grows trees using a “leaf–wise” strategy instead of the more common “level–wise” approach. This means that instead of expanding all nodes at the same depth, the algorithm picks the leaf with the highest potential gain and splits it first. This targeted growth allows the model to find deeper, more informative partitions where they matter most. The resulting trees may be more complex, so LightGBM includes built–in regularization to control depth and prevent overfitting.

Like XGBoost, LightGBM naturally handles nonlinearities, interactions, and mixed–scale predictors. Its efficiency makes it well suited for repeated tuning or evaluation in comparative studies. In population–health applications, LightGBM has the practical advantage of training quickly even when using many socioeconomic, behavioral, and contextual predictors.

\subsection{Deep Neural Network (Multilayer Perceptron)}

A multilayer perceptron (MLP) is a Deep Neural Network approach responsible for understanding the associations between linear and non-linear data sets. It is a kind of feed-forward neural network that may be separated into three different layers called the input layer, output layer, and hidden layer. The signal that will be processed is transferred to the input layer. It is the output layer's responsibility to do the appropriate tasks. The actual computational engine of the MLP consists of an arbitrary number of hidden layers located between the input and output layers. An MLP functions in a way comparable to a feed-forward network. Data moves forward from the input layer to the output layer. 






\subsection{TabNet}






TabNet\cite{tabnet} is a deep–learning architecture designed specifically for tabular data, where classical neural networks often struggle. Instead of treating every feature the same way at every layer, TabNet introduces an attention mechanism that allows the model to decide which features to focus on at each decision step. The goal is to blend the flexibility of deep learning with the structured, feature–wise reasoning that drives tree–based models.

A TabNet model processes data through a sequence of decision blocks. At each step, the network selects a subset of features using an attention mask. This mask is not fixed; it is learned during training. The selected features flow through a small network that extracts meaningful representations, while unselected features are suppressed for that step. Each step contributes part of the final prediction, and the model combines what it learns across all steps.

One benefit of this design is interpretability. Because the model generates explicit feature masks, it is possible to see which variables influenced decisions at different stages. This provides a more transparent deep learning approach for domains, such as population health, where understanding feature importance is part of the analytic goal.

TabNet also uses sparse regularization. This encourages the model to focus on a small number of informative features at each step rather than spreading attention across everything. Sparse feature selection helps the network avoid overfitting and mimics the behavior of boosting models that refine predictions by concentrating on the most relevant predictors.

In practice, TabNet offers a middle ground: it can capture complex nonlinear patterns like a deep neural network, but it also provides structured interpretability similar to tree–based methods. Its design makes it suitable for heterogeneous survey data, where predictors vary in scale and type.

\section{Experiments}
\label{experiment}

Here we apply our proposed method to the models we describe in Section \ref{Methodology} and show our findings.
    
\subsection{Dataset}

We use data from the 2021 National Survey of Children’s Health (NSCH)\cite{nsch}, an annual, nationally representative survey administered by the U.S. Census Bureau. The NSCH collects detailed information on the physical, emotional, and behavioral health of children ages 0–17, along with extensive measures of family environments, household socioeconomic conditions, and neighborhood characteristics. 

The dataset provides parent-reported information on child health status, health behaviors, healthcare access, adverse childhood experiences, parental functioning, and community environments. For this analysis, the focus is on children aged 10–17 years, a developmental period in which independent behaviors, psychosocial stressors, and environmental exposures play increasing roles in shaping weight outcomes. The analytic sample includes all respondents within this age range who supply complete information for the outcome variable and the selected predictors. Descriptive characteristics of the analytic sample, including demographic composition, behavioral indicators, household socioeconomic conditions, and community context, are summarized in Table~\ref{tab:descriptive}.

The dataset contains several categories of variables relevant to childhood obesity research:

\begin{description}

\item[1. Behavioral Variables:]
The survey includes detailed measures of diet, activity, and daily routines. For this study, the key behavioral indicators include daily physical activity (\texttt{PhysAct\_21}), screen-time duration (\texttt{ScreenTime\_21}), sugar-sweetened beverage intake (\texttt{SugarDrink\_21}), vegetable and fruit consumption (\texttt{vegetables\_21}, \texttt{fruit\_21}), sleep duration (\texttt{HrsSleep\_21}), and weight-related concern (\texttt{WgtConcn\_21}). These measures capture modifiable behaviors linked to energy balance.

\item[2. Household and Family Variables:]
The dataset provides indicators of socioeconomic position, family structure, parental education (\texttt{AdultEduc\_21}), household income relative to the federal poverty level (\texttt{povlev4\_21}), parental stress or aggravation (\texttt{ParAggrav\_21}), emotional support (\texttt{EmSupport\_21}), coping ability (\texttt{ParCoping\_21}), smoking exposure (\texttt{smoking\_21}, \texttt{SmkInside\_21}), food insecurity (\texttt{FoodSit\_21}), and adverse childhood experiences (\texttt{ACEincome\_21}, \texttt{ACEdomviol\_21}, \texttt{ACEdrug\_21}, \texttt{ACEneighviol\_21}, \texttt{ACEdiscrim\_21}). These variables reflect structural and psychosocial conditions that influence obesity risk.

\item[3. Community and Neighborhood Variables:]
Neighborhood support (\texttt{NbhdSupp\_21}), safety (\texttt{NbhdSafe\_21}, \texttt{SchlSafe\_21}), environmental detractors such as litter or vandalism (\texttt{NbhdDetract\_21}), and the availability of amenities such as parks or recreation centers (\texttt{NbhdAmenities\_21}) provide contextual markers of community health. Metropolitan vs. non-metropolitan residence (\texttt{METRO\_YN}) and core-based statistical area designation (\texttt{CBSAFP\_YN}) capture geographic variation.

\item[4. Demographic Variables:]
Age (\texttt{SC\_AGE\_YEARS}), sex (\texttt{SC\_SEX}), and race/ethnicity (\texttt{SC\_RACE\_R}) describe population heterogeneity.
\end{description}

The outcome variable, overweight/obesity, is derived from \texttt{BMI3\_21}, which classifies adolescents as underweight, normal weight, or overweight/obese. In this study, we use a binary indicator distinguishing overweight/obese from normal weight. Underweight cases are removed to avoid uninformative heterogeneity.

\begin{figure}[!ht]
    \centering
    \includegraphics[width=0.85\textwidth]{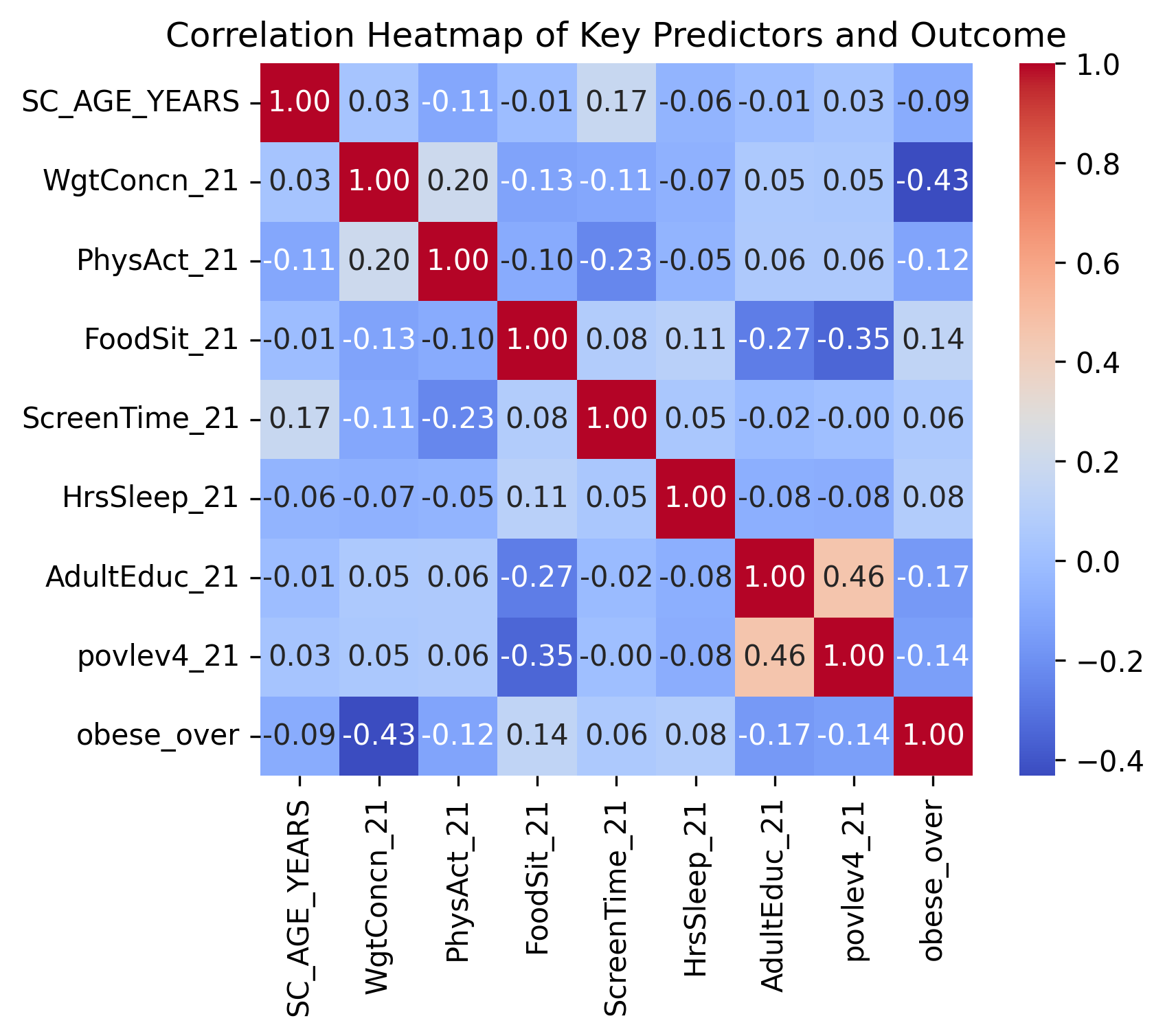}
    \caption{Correlation Heatmap of Key Predictors and Outcome.
    }
    \label{fig:correlation}
\end{figure}

To assess relationships among key behavioral, socioeconomic, and demographic predictors prior to model fitting, we examine correlations between the selected variables and the outcome variable. Figure~\ref{fig:correlation} presents a correlation heatmap summarizing the direction and magnitude of linear associations.

The NSCH dataset is well suited for predictive modeling because it captures heterogeneous, real-world conditions across diverse U.S. populations. Its breadth allows models to incorporate individual, familial, and neighborhood-level determinants simultaneously, and its national sampling design supports generalizable inferences about childhood obesity patterns in the United States.


\begin{table}[!ht]
\centering
\caption{Descriptive characteristics of the analytic sample (NSCH 2021, ages 10-17)}
\label{tab:descriptive}
\begin{tabular}{ll}
\toprule
\textbf{Characteristic} & \textbf{Value} \\
\midrule
\multicolumn{2}{l}{\textit{Sample size}} \\
Total participants & 18,792 \\
\midrule
\multicolumn{2}{l}{\textit{Outcome prevalence}} \\
Normal weight (\%) & 67.6 \\
Overweight / Obese (\%) & 32.4 \\
\midrule
\multicolumn{2}{l}{\textit{Demographics}} \\
Mean age, years (SD) & 13.80 (2.28) \\
Male (\%) & 51.4 \\
Female (\%) & 48.6 \\
\midrule
\multicolumn{2}{l}{\textit{Race / ethnicity (\%)}} \\
Hispanic & 76.7 \\
White, non-Hispanic & 7.5 \\
Black, non-Hispanic & 1.3 \\
Asian, non-Hispanic & 5.7 \\
American Indian / Alaska Native, non-Hispanic & 0.9 \\
Multiracial, non-Hispanic & 7.9 \\
\midrule
\multicolumn{2}{l}{\textit{Behavioral characteristics}} \\
Physical activity frequency (mean, SD) & 2.50 (0.91) \\
Screen time (mean, SD) & 3.84 (1.12) \\
\midrule
\multicolumn{2}{l}{\textit{Community characteristics (\%)}} \\
Neighborhood safe -- definitely agree & 72.0 \\
Neighborhood safe -- somewhat agree & 24.9 \\
Neighborhood safe -- disagree & 3.1 \\
Supportive neighborhood (yes) & 60.9 \\
Supportive neighborhood (no) & 39.1 \\
Metropolitan residence & 80.9 \\
Non-metropolitan residence & 19.1 \\
\midrule
\multicolumn{2}{l}{\textit{Household poverty level (\% of Federal Poverty Level [FPL])}} \\
$\geq$400\% FPL & 40.5 \\
200--399\% FPL & 30.4 \\
100--199\% FPL & 16.2 \\
$<$100\% FPL & 12.8 \\
\bottomrule
\end{tabular}
\end{table}

\subsection{Experimental Setup} 

We use one testbed device to run all experiments. The device consists of an AMD Ryzen 9~8945HS CPU, an NVIDIA RTX~4070 GPU, and 32~GB RAM. All analyses were conducted in Python (version 3.9). Machine-learning and statistical modeling leverage PyTorch (v2.0), TorchMetrics (v0.11.4), scikit-learn (v1.5.2), XGBoost (v2.1.4), and LightGBM (v4.6.0). Data manipulation and preprocessing are performed using pandas (v2.2.3), NumPy (v1.26.3), and SciPy (v1.13.1). Visualization and exploratory analysis relied on Matplotlib (v3.9.2) and Seaborn (v0.13.2). Tabular deep-learning models are implemented using pytorch-tabnet (v4.1.0). All models are trained under identical conditions to ensure comparability. We fix all random seeds across NumPy, PyTorch, and Python’s \texttt{random} module so that repeated runs produce the same results.



\subsubsection{Data Pre-Processing}
Before model training, we initially prepare the dataset through a standardized preprocessing pipeline. Initially, we select the predictor variables and the output variable. Then, we select the age from 10 to 17. Variables with survey–specific codes such as “missing” are recoded as missing values and represented with NaN. Continuous predictors are standardized to zero mean and unit variance using statistics computed from the training data. We impute missing values in all predictors so that every model receives the same completed dataset. Categorical predictors are one–hot encoded to avoid imposing an ordinal structure. 

The dataset is split into training, validation, and testing partitions. The test split remains untouched throughout all experiments and is used only for final evaluation. Within the training split, models that support validation monitoring use an the validation split for early stopping.

\subsection{Model Training Procedures}


All models are trained using supervised learning on the same training partition to ensure fair comparison. No ensemble stacking or blending is performed; each model is trained independently and evaluated on the same test set.

Logistic regression is trained using maximum likelihood estimation with an $\ell_2$ regularization penalty and a fixed maximum number of iterations to ensure convergence. Predictor variables are standardized prior to training to place coefficients on comparable scales.

Random Forest and Gradient Boosting models are trained using decision trees as base learners. Random Forest uses a large number of trees to reduce variance through bagging, while Gradient Boosting sequentially fits trees to correct residual errors. Default impurity-based splitting criteria are used. We do not apply any explicit class weighting, allowing observed class imbalance to reflect the natural prevalence of overweight and obesity.

XGBoost and LightGBM are trained using gradient boosting with shrinkage and subsampling. Learning rates, tree depth, and number of estimators are fixed across runs to balance bias and variance. Early stopping is not used for these models; instead, model capacity is controlled through conservative hyperparameter choices.

The deep neural network model is trained using mini–batch gradient descent with the Adam optimizer. Binary cross–entropy loss is used. 
Training proceeds for a maximum number of epochs with early stopping based on validation AUC to prevent overfitting. Also, it incorporates batch normalization, dropout regularization, and learning-rate decay.

TabNet is trained using its native attentive learning procedure. Feature selection masks are learned jointly with prediction through sparse attention mechanisms. We apply early stopping using validation AUC. No additional class rebalancing is performed beyond TabNet’s internal optimization.

Across all models, hyperparameters are selected to reflect commonly used configurations rather than aggressive tuning, emphasizing comparability and generalizability over maximal performance.


\subsection{Metrics}

Model performance is evaluated using a combination of discrimination, classification, and calibration metrics to reflect priorities in public health research.

Discrimination is assessed using the area under the receiver operating characteristic curve (AUC). AUC measures a model’s ability to rank overweight or obese children higher than normal–weight children across all classification thresholds. 

Classification performance is summarized using accuracy, precision, recall, and F1 score at a fixed probability threshold. Accuracy reflects overall correctness but is sensitive to class imbalance. Precision quantifies the proportion of predicted high–risk children who are truly overweight or obese, which is relevant when interventions are resource–constrained. Recall measures the proportion of overweight or obese children correctly identified, reflecting sensitivity.
The F1 score balances precision and recall, providing a single summary of classification trade–offs.

Calibration is evaluated using the Brier score, which measures the mean squared difference between predicted probabilities and observed outcomes. Lower Brier scores indicate better probabilistic calibration, which is an important property when predicted risks may inform screening or policy decisions.

\subsection{Interpretability Procedures}

We conduct interpretability analyses for logistic regression, LightGBM, and TabNet to identify predictors most strongly associated with overweight and obesity. We select these models because they provide either intrinsic interpretability or well-established global explanation mechanisms \cite{Molnar2025-ja}, enabling transparent assessment of predictor contributions.

For logistic regression, we derive interpretability directly from model coefficients, which quantify the direction and relative strength of associations between predictors and the outcome under a linear modeling assumption. We examine coefficient magnitudes to identify variables most strongly associated with obesity risk.

For LightGBM, we assess global feature importance using permutation-based importance measures, which quantify the decrease in model performance when individual predictors are randomly permuted. This approach captures the contribution of each variable to predictive accuracy while accommodating nonlinear effects and interactions learned by the gradient-boosted trees.

For TabNet, we leverage its native attentive learning mechanism to obtain built-in interpretability. During training, TabNet learns sparse attention masks at each decision step that indicate the relative contribution of individual predictors to the model’s predictions. 
We aggregate TabNet’s learned feature importance scores across one-hot encoded categories to produce global, variable-level importance estimates that reflect how frequently and strongly each variable is used across the model’s decision process.

We report interpretability results descriptively to contextualize model behavior and highlight influential behavioral, household, and community-level factors rather than to infer causal relationships. Because interpretability mechanisms differ fundamentally across model classes, we do not directly compare explanation outputs across models but instead interpret each set of results within the context of its underlying modeling framework.

\section{Results}
\label{sec:results}
In this section, we describe the findings after successfully experimenting with our proposed model under different scenarios.


\subsection{Overall Predictive Performance}

\begin{figure}[!ht]
\centering
\begin{subfigure}{0.42\textwidth}
    \centering
    \includegraphics[width=\textwidth]{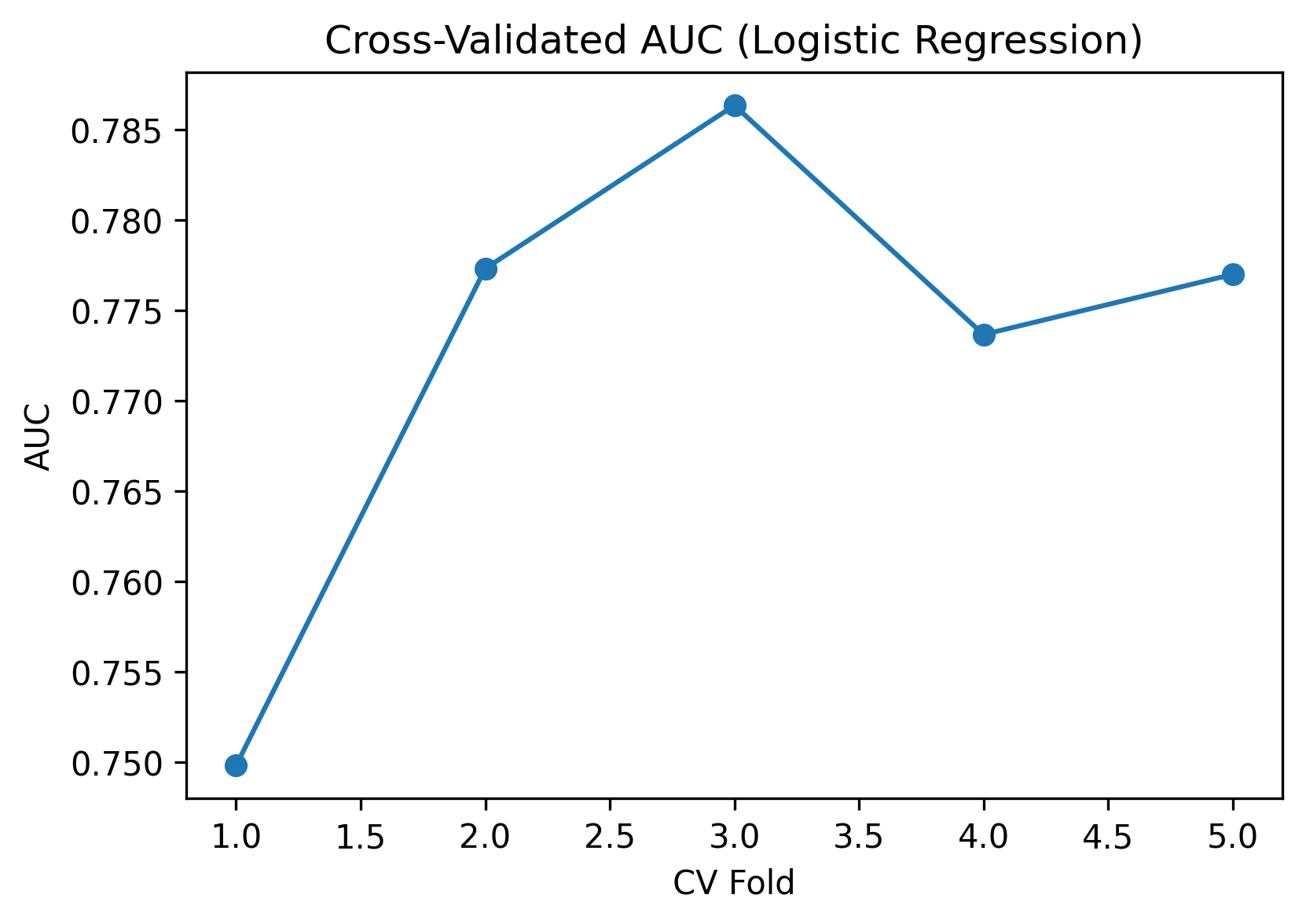}
    \caption{Cross-validated AUC across folds}
    \label{fig:cv_auc}
\end{subfigure}
\hfill
\begin{subfigure}{0.42\textwidth}
    \centering
    \includegraphics[width=\textwidth]{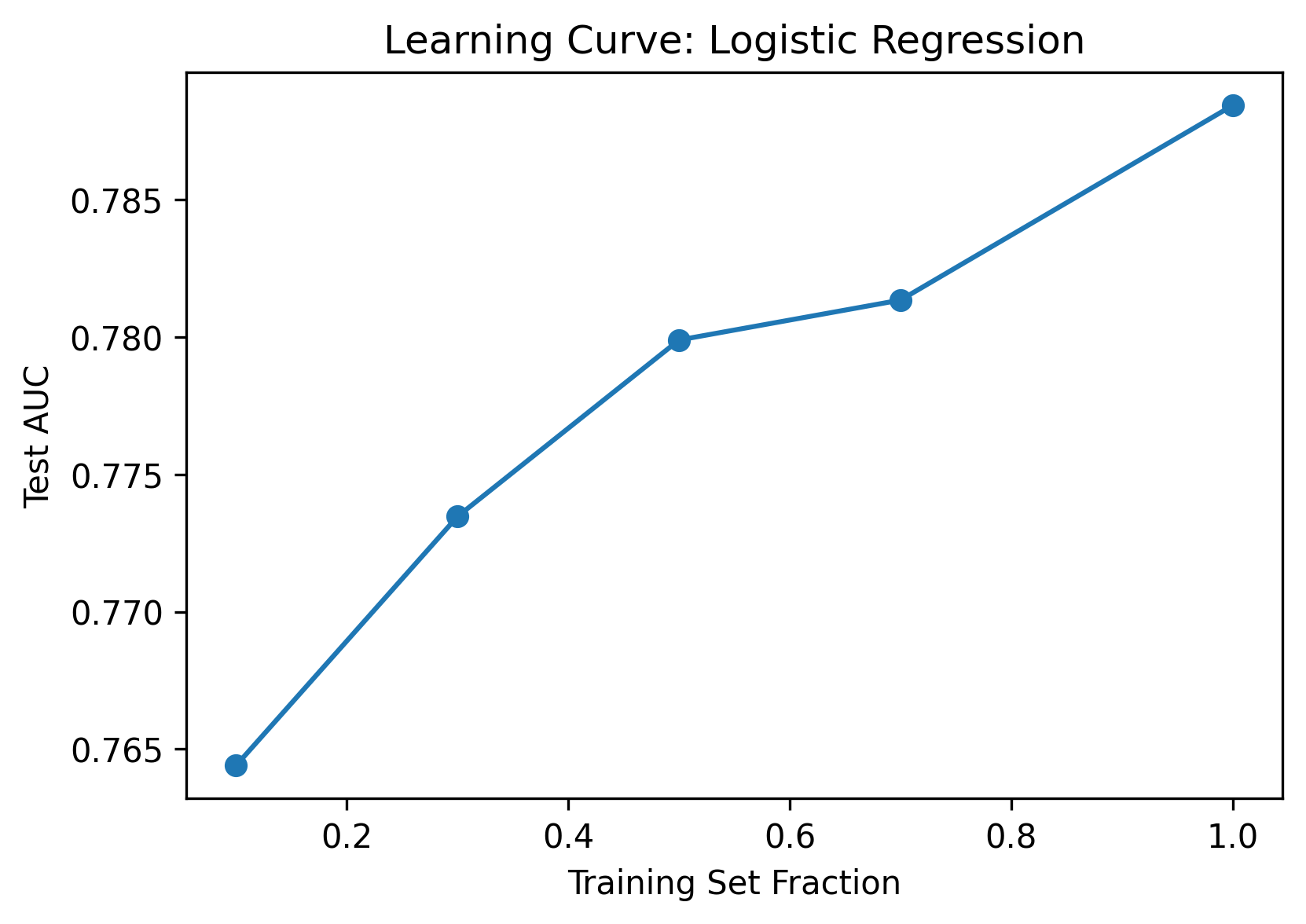}
    \caption{Learning curve showing AUC versus training size}
    \label{fig:learning_curve}
\end{subfigure}
\caption{Model stability assessment using cross-validation and learning curves for logistic regression.}
\label{fig:model_stability}
\end{figure}

Logistic regression achieves the strongest overall discrimination among all models (AUC = 0.788), with high accuracy (0.769) and precision (0.830), but recall remains low (0.360). This pattern reflects conservative classification behavior, producing reliable positive predictions while substantially under-identifying overweight and obese adolescents. Calibration is relatively strong (Brier = 0.160), supporting the continued utility of logistic regression for population-level risk estimation.
To assess model stability, we perform cross-validation and learning-curve analysis for logistic regression. Results from Figure \ref{fig:model_stability} show that cross-validated discrimination remains stable across folds, with a mean AUC of 0.773 and low variability (SD = 0.012), indicating robust and consistent performance across resampled training sets. Learning-curve analysis shows smooth convergence of training and validation performance as sample size increases, with no evidence of severe overfitting or instability. Together, these results suggest that logistic regression provides reliable and well-calibrated baseline performance in this nationally representative dataset, supporting its use as a stable reference model for comparative evaluation.

Random Forest exhibits slightly lower discrimination (AUC = 0.763) with comparable accuracy (0.765) and precision (0.822). Recall remains low (0.350), and overall performance closely mirrors that of logistic regression, suggesting limited incremental benefit from nonlinear tree aggregation in this setting.

Gradient Boosting provides similarly strong performance, achieving high discrimination (AUC = 0.786) and stable accuracy (0.765). Precision remains high (0.822), calibration improves modestly (Brier = 0.161), and recall remains constrained (0.350). These results indicate improved probability estimation without substantive gains in sensitivity.

XGBoost and LightGBM demonstrate comparable performance profiles. XGBoost achieves an AUC of 0.768 with accuracy of 0.761, while LightGBM attains a slightly higher AUC of 0.772 with accuracy of 0.768. Both models improve recall modestly (0.384 for each), resulting in higher F1 scores relative to logistic regression and Random Forest, while calibration remains stable. These gains reflect improved sensitivity at the expense of reduced precision.

The deep learning model exhibits a well balanced performance across metrics. Discrimination (AUC = 0.779) and accuracy (0.766) remain comparable to boosting-based approaches, while recall increases to 0.397, yielding the highest F1 score (0.524) among all models. Calibration remains acceptable (Brier = 0.162). Unlike earlier iterations, this model does not disproportionately prioritize sensitivity and instead converges toward ensemble-based performance.

TabNet underperforms all other approaches. Discrimination is substantially lower (AUC = 0.655), recall is poor (0.264), and calibration deteriorates noticably (Brier = 0.194). These results suggest that sparse attention-based architectures are poorly matched to the structure and coding conventions of NSCH survey data.

Overall, no model demonstrates uniformly superior performance. Table~\ref{tab:performance} summarizes the overall performance of all models.

\begin{table}[ht]
\centering
\caption{Predictive performance of all models on the test set. Best values are bolded.}
\label{tab:performance}
\begin{tabular}{lcccccc}
\hline
Model & AUC & Accuracy & Precision & Recall & F1 & Brier \\
\hline
Logistic Regression 
& \textbf{0.788} & \textbf{0.769} & \textbf{0.830} & 0.360 & 0.502 & \textbf{0.160} \\

Random Forest 
& 0.763 & 0.765 & 0.822 & 0.350 & 0.491 & 0.166 \\

Gradient Boosting 
& 0.786 & 0.765 & 0.822 & 0.350 & 0.491 & 0.161 \\

XGBoost 
& 0.768 & 0.761 & 0.760 & 0.384 & 0.510 & 0.166 \\

LightGBM 
& 0.772 & 0.768 & 0.793 & 0.384 & 0.518 & 0.164 \\

Deep Learning (MLP) 
& 0.779 & 0.766 & 0.771 & \textbf{0.397} & \textbf{0.524} & 0.162 \\

TabNet 
& 0.655 & 0.736 & 0.767 & 0.264 & 0.393 & 0.194 \\
\hline
\end{tabular}
\end{table}



\subsection{Feature Importance and Associated Factors}

Feature-importance analyses reveal consistent multilevel patterns across modeling approaches, despite differences in representation and interpretability mechanisms.

Logistic regression coefficients identify parental concern about child weight as the strongest predictor of overweight and obesity, followed by demographic characteristics such as race, sex, and household educational attainment. Behavioral factors, including physical activity frequency, and household socioeconomic indicators, such as poverty level, food insufficiency, and household smoking exposure, also exhibit meaningful associations. These effects emerge at the level of specific response categories, reflecting interpretable but linear relationships between predictors and obesity risk.

Permutation-based importance from LightGBM reinforces the dominant role of parental weight concern, while also highlighting adult educational attainment, child age, sex, and race as influential predictors. Behavioral variables such as sleep duration, screen time, and physical activity contribute modestly, alongside indicators of neighborhood support and household food security. These results suggest sensitivity to nonlinear interactions among demographic, behavioral, and contextual factors, while largely corroborating the predictors identified by the linear model.

Although TabNet underperforms in predictive accuracy, its aggregated attention-based feature importance provides descriptive insight into domain-level emphasis. Parental weight concern again emerges as the most influential variable, followed by race, physical activity, screen time, neighborhood amenities and detracting elements, household education, family size, and poverty-related indicators. Adverse childhood experiences and neighborhood safety measures also receive moderate attention, emphasizing the combined influence of household stressors and environmental context.

Taken together, feature-importance results consistently indicate that adolescent overweight and obesity are shaped by an interplay of behavioral factors, family socioeconomic conditions, parental perceptions, and neighborhood environments. Nonlinear and attention-based models amplify the prominence of contextual and environmental variables but do not fundamentally alter the core patterns observed in logistic regression, reinforcing a multilevel structure of obesity risk rather than identifying distinct or model-specific drivers.


\begin{table}[ht]
\centering
\caption{Top 10 influential features by model. Values in parentheses indicate importance magnitude: absolute standardized coefficients for Logistic Regression, permutation importance for LightGBM, and aggregated attention importance for TabNet. Bolded features appear consistently across all three models.}
\label{tab:feature_importance_models}
\begin{tabular}{lll}
\hline
\textbf{Logistic Regression} & \textbf{LightGBM} & \textbf{TabNet} \\
\hline
\textbf{WgtConcn\_21} (2.894) & \textbf{WgtConcn\_21} (0.117) & \textbf{WgtConcn\_21} (0.111) \\

\textbf{SC\_RACE\_R} (0.433) & \textbf{AdultEduc\_21} (0.024) & \textbf{SC\_RACE\_R} (0.082) \\

\textbf{AdultEduc\_21} (0.382) & SC\_AGE\_YEARS (0.013) & \textbf{PhysAct\_21} (0.052) \\

\textbf{SC\_SEX} (0.248) & \textbf{SC\_SEX} (0.006) & ScreenTime\_21 (0.052) \\

PhysAct\_21 (0.181) & SC\_RACE\_R (0.005) & NbhdAmenities\_21 (0.049) \\

povlev4\_21 (0.176) & NbhdSupp\_21 (0.003) & NbhdDetract\_21 (0.047) \\

ParCoping\_21 (0.159) & FoodSit\_21 (0.002) & \textbf{AdultEduc\_21} (0.046) \\

FoodSit\_21 (0.142) & HrsSleep\_21 (0.002) & FamCount\_21 (0.043) \\

SmkInside\_21 (0.127) & ScreenTime\_21 (0.002) & povlev4\_21 (0.041) \\

FamCount\_21 (0.106) & PhysAct\_21 (0.001) & ACEincome\_21 (0.039) \\
\hline
\end{tabular}
\end{table}

\begin{figure}[htbp]
\centering
\begin{subfigure}{0.32\textwidth}
    \centering
    \includegraphics[width=\textwidth]{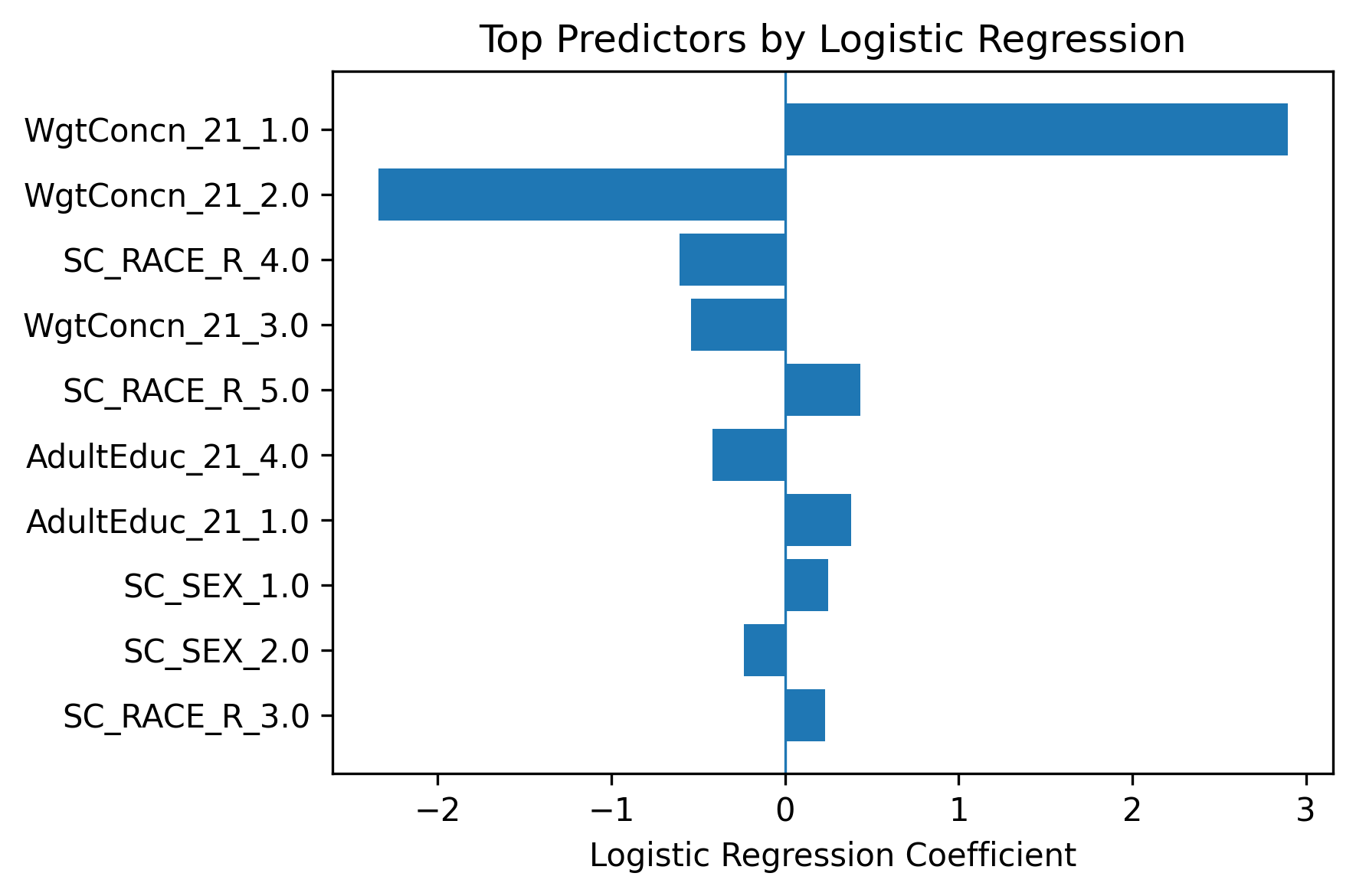}
    \caption{Logistic regression feature importance}
    \label{fig:logit_imp}
\end{subfigure}
\hfill
\begin{subfigure}{0.32\textwidth}
    \centering
    \includegraphics[width=\textwidth]{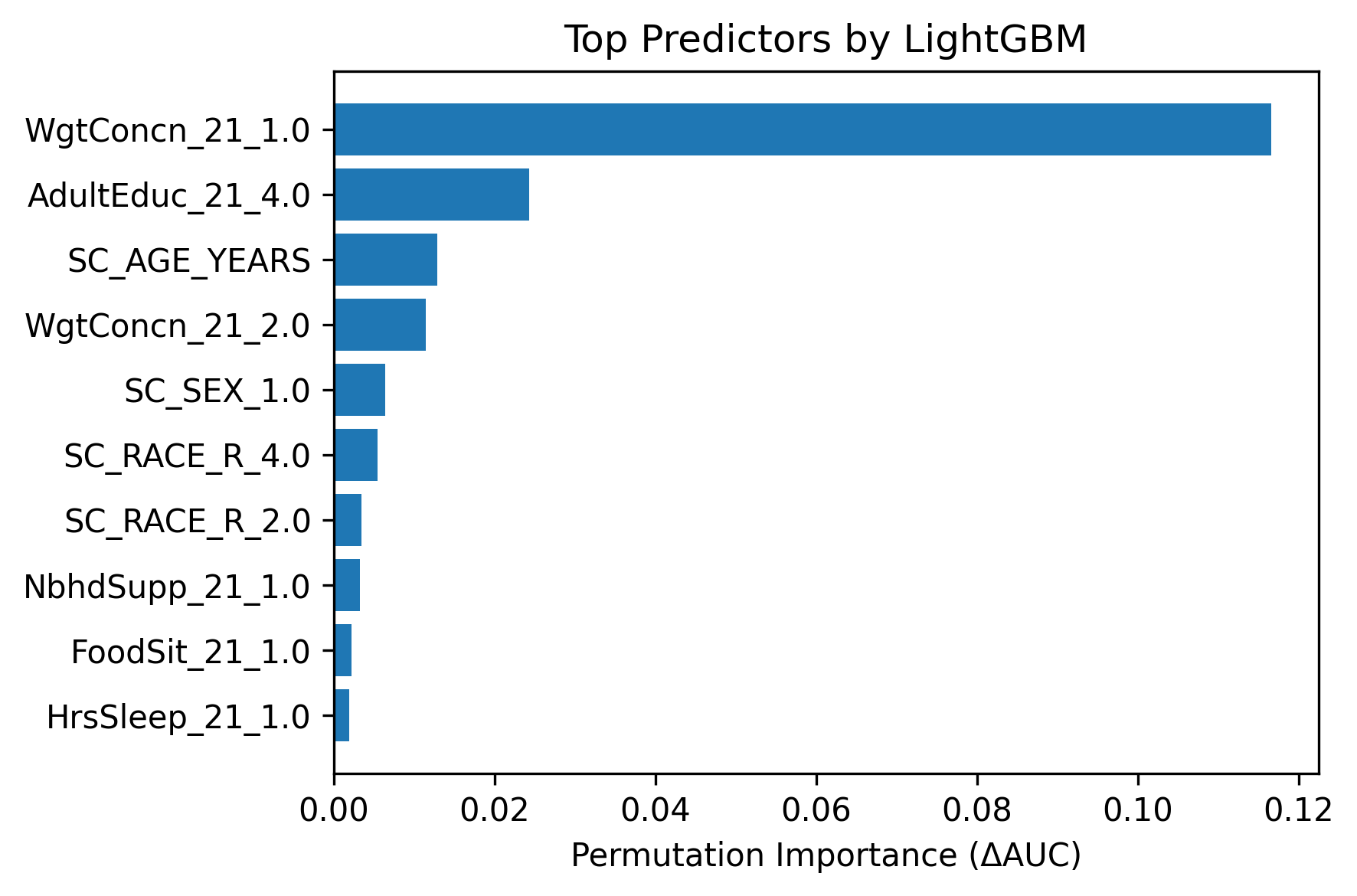}
    \caption{LightGBM permutation importance}
    \label{fig:lgbm_imp}
\end{subfigure}
\hfill
\begin{subfigure}{0.32\textwidth}
    \centering
    \includegraphics[width=\textwidth]{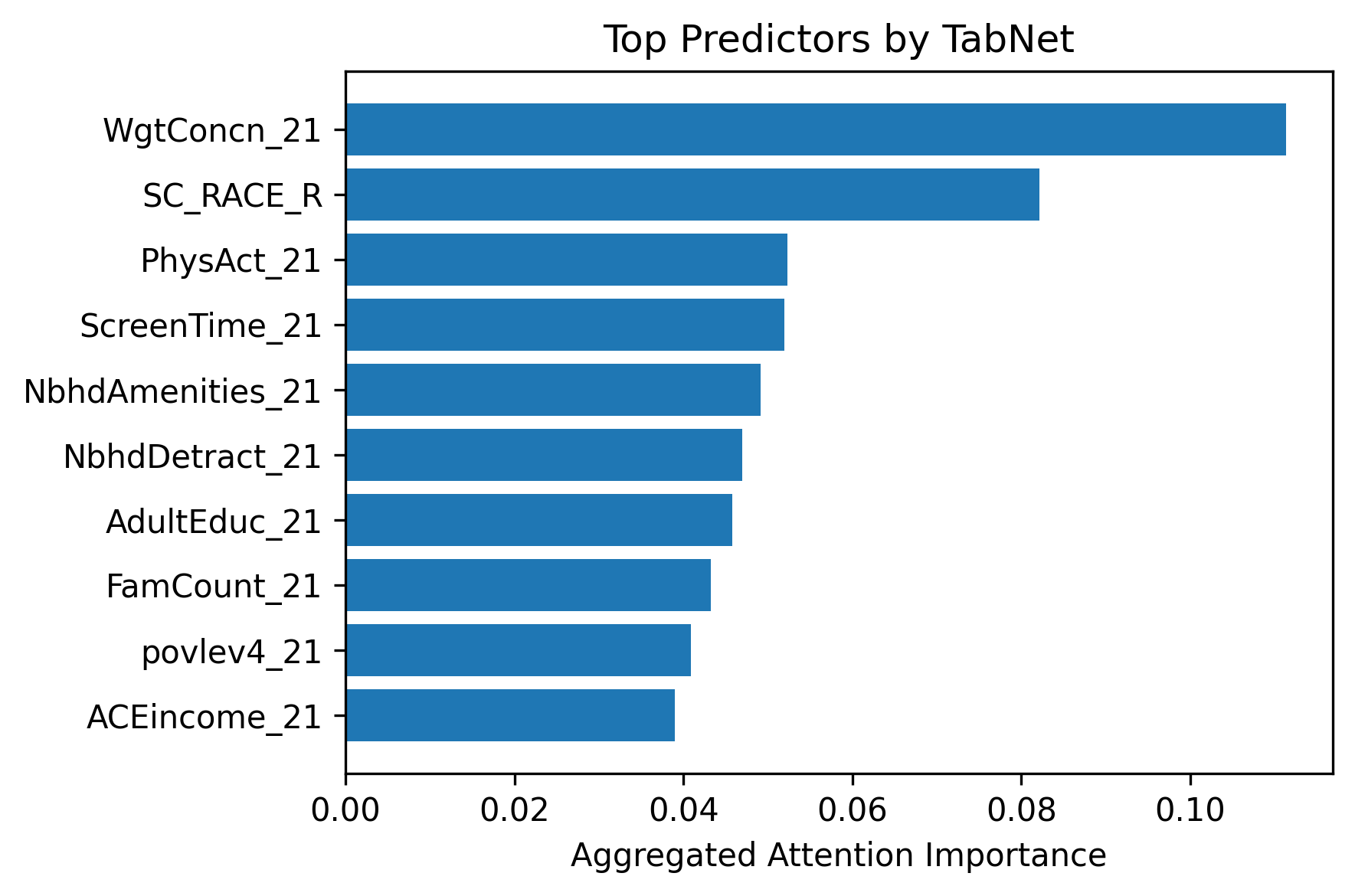}
    \caption{TabNet aggregated feature importance}
    \label{fig:tabnet_imp}
\end{subfigure}

\caption{Top predictors of adolescent overweight and obesity across modeling approaches.}
\label{fig:three_model_importance}
\end{figure}

\section{Fairness Evaluation}
\label{sec:fairness_evaluation}

This section examines group-conditional model performance across sex, race, poverty level, and metropolitan status. Performance is evaluated using discrimination (AUC), classification behavior (precision, recall, and F1 score), and calibration (Brier score). The analysis is descriptive and does not impose explicit fairness constraints or reweighting strategies.

\textbf{Sex.}
Sex-based differences are modest and consistent across modeling approaches. Discrimination remains comparable between groups for logistic regression, ensemble models, boosting methods, and deep learning, indicating similar ranking ability across sexes. Group~1 (Male) generally exhibits higher recall and F1 scores, reflecting greater sensitivity to overweight and obesity, while Group~2 (Female) consistently shows higher accuracy, precision, and better calibration, indicating more conservative classification behavior. These trade-offs persist across models and suggest that sex-related differences primarily reflect sensitivity–specificity balance rather than systematic disparities in discrimination. TabNet performs poorly for both groups, with low recall and weak calibration, reflecting overall model instability rather than sex-specific bias.

\textbf{Race.}
Race-based disparities are pronounced and persistent across all models. Substantial variation is observed in discrimination, recall, and calibration across racial groups, with no model achieving uniformly stable performance. Several groups exhibit relatively high recall but degraded calibration, while others show low sensitivity and better-calibrated probabilities, indicating uneven error distribution. These patterns appear consistently across logistic regression, ensemble methods, boosting models, and deep learning, suggesting that increased model complexity does not resolve race-related performance gaps. TabNet demonstrates particularly unstable behavior, with low discrimination, highly variable recall, and poor calibration across multiple racial groups, especially those with smaller sample sizes.

\textbf{Poverty level.}
Model performance varies systematically across socioeconomic strata. Higher-income groups tend to achieve stronger discrimination and better calibration, whereas lower-income groups often exhibit higher recall accompanied by worse Brier scores and reduced precision. 
Across all models, better detection of overweight and obesity in lower-income groups comes with less stable probability estimates, meaning the predictions are more uncertain. This trade-off appears in traditional models, tree-based methods, and deep-learning models alike, indicating that income-related differences make prediction less consistent. In higher-income groups, predictions are more precise, but many true cases are still missed, resulting in consistent under-identification of overweight and obesity in these populations.

\textbf{Metro vs.\ rural.}
Geographic disparities are smaller than those observed for race and poverty but remain consistent across models. Metropolitan groups generally achieve slightly higher discrimination and recall, while rural groups exhibit reduced sensitivity and marginally worse calibration. These differences are observed across all modeling approaches and likely reflect contextual variation, access-related factors, and data sparsity rather than algorithm-specific bias.

\textbf{Cross-model patterns.}
Across all subgroup analyses, logistic regression and tree-based models demonstrate stable but conservative behavior, characterized by high precision and suppressed recall. Boosting models modestly improve sensitivity while maintaining comparable calibration. Deep-learning models achieve the highest recall and F1 scores across many subgroups but do not eliminate disparities and occasionally introduce mild calibration degradation. TabNet consistently underperforms across nearly all subgroup dimensions, particularly for race and poverty, and fails to provide reliable or equitable predictions in this survey-based setting.

Overall, subgroup disparities persist across all algorithms, indicating that fairness limitations are driven primarily by structural characteristics of the data, population heterogeneity, and unequal subgroup representation rather than model choice alone.

\begin{table}[!ht]
\centering
\scriptsize
\begin{minipage}{0.48\linewidth}
\centering
\caption{Fairness by sex (group-conditional performance).}
\label{tab:fairness_sex}
\begin{tabular}{llcccccc}
\hline
Model & Group & AUC & Precision & Recall & F1 & Brier \\
\hline
Logistic Regression & 1 & 0.789 & 0.830 & 0.380 & 0.521 & 0.171 \\
Logistic Regression & 2 & 0.774 & 0.828 & 0.331 & 0.473 & 0.149 \\

Random Forest & 1 & 0.758 & 0.841 & 0.366 & 0.510 & 0.178 \\
Random Forest & 2 & 0.760 & 0.800 & 0.323 & 0.460 & 0.154 \\

Gradient Boosting & 1 & 0.787 & 0.830 & 0.366 & 0.508 & 0.171 \\
Gradient Boosting & 2 & 0.771 & 0.810 & 0.327 & 0.466 & 0.151 \\

XGBoost & 1 & 0.770 & 0.771 & 0.402 & 0.528 & 0.176 \\
XGBoost & 2 & 0.754 & 0.742 & 0.359 & 0.484 & 0.155 \\

LightGBM & 1 & 0.781 & 0.806 & 0.402 & 0.536 & 0.172 \\
LightGBM & 2 & 0.750 & 0.774 & 0.359 & 0.490 & 0.155 \\

Deep Learning & 1 & 0.776 & 0.785 & 0.424 & 0.550 & 0.172 \\
Deep Learning & 2 & 0.766 & 0.748 & 0.359 & 0.485 & 0.152 \\

TabNet & 1 & 0.650 & 0.792 & 0.274 & 0.407 & 0.208 \\
TabNet & 2 & 0.656 & 0.729 & 0.250 & 0.372 & 0.179 \\
\hline
\end{tabular}
\end{minipage}
\hfill
\begin{minipage}{0.48\linewidth}
\centering
\caption{Fairness by metro vs.\ rural (group-conditional performance).}
\label{tab:fairness_metro}
\begin{tabular}{llcccccc}
\hline
Model & Group & AUC & Precision & Recall & F1 & Brier \\
\hline
Logistic Regression & 1 & 0.793 & 0.818 & 0.373 & 0.513 & 0.157 \\
Logistic Regression & 2 & 0.779 & 0.870 & 0.320 & 0.468 & 0.177 \\

Random Forest & 1 & 0.767 & 0.817 & 0.371 & 0.510 & 0.163 \\
Random Forest & 2 & 0.754 & 0.850 & 0.272 & 0.412 & 0.184 \\

Gradient Boosting & 1 & 0.794 & 0.814 & 0.373 & 0.512 & 0.157 \\
Gradient Boosting & 2 & 0.770 & 0.872 & 0.272 & 0.415 & 0.180 \\

XGBoost & 1 & 0.782 & 0.763 & 0.408 & 0.532 & 0.160 \\
XGBoost & 2 & 0.731 & 0.780 & 0.312 & 0.446 & 0.189 \\

LightGBM & 1 & 0.785 & 0.786 & 0.406 & 0.535 & 0.158 \\
LightGBM & 2 & 0.746 & 0.848 & 0.312 & 0.456 & 0.185 \\

Deep Learning & 1 & 0.786 & 0.770 & 0.417 & 0.541 & 0.158 \\
Deep Learning & 2 & 0.752 & 0.764 & 0.336 & 0.467 & 0.183 \\

TabNet & 1 & 0.660 & 0.784 & 0.267 & 0.399 & 0.190 \\
TabNet & 2 & 0.616 & 0.688 & 0.264 & 0.382 & 0.218 \\
\hline
\end{tabular}
\end{minipage}
\end{table}


\begin{table}[!ht]
\centering
\scriptsize
\begin{minipage}{0.48\linewidth}
\centering
\caption{Fairness by race (group-conditional performance).}
\label{tab:fairness_race}
\begin{tabular}{llcccccc}
\hline
Model & Group & AUC & Precision & Recall & F1 & Brier \\
\hline
Logistic Regression & 1 & 0.791 & 0.837 & 0.337 & 0.481 & 0.156 \\
Logistic Regression & 2 & 0.735 & 0.810 & 0.500 & 0.618 & 0.201 \\
Logistic Regression & 3 & 0.768 & 0.833 & 0.714 & 0.769 & 0.155 \\
Logistic Regression & 4 & 0.810 & 0.818 & 0.346 & 0.486 & 0.132 \\
Logistic Regression & 5 & 0.736 & 0.833 & 0.417 & 0.556 & 0.217 \\
Logistic Regression & 7 & 0.751 & 0.810 & 0.315 & 0.453 & 0.172 \\

Random Forest & 1 & 0.759 & 0.805 & 0.337 & 0.475 & 0.164 \\
Random Forest & 2 & 0.725 & 0.875 & 0.412 & 0.560 & 0.206 \\
Random Forest & 3 & 0.902 & 1.000 & 0.714 & 0.833 & 0.139 \\
Random Forest & 4 & 0.802 & 0.818 & 0.346 & 0.486 & 0.135 \\
Random Forest & 5 & 0.708 & 0.800 & 0.333 & 0.471 & 0.233 \\
Random Forest & 7 & 0.753 & 0.895 & 0.315 & 0.466 & 0.174 \\

Gradient Boosting & 1 & 0.786 & 0.826 & 0.333 & 0.474 & 0.157 \\
Gradient Boosting & 2 & 0.737 & 0.821 & 0.471 & 0.598 & 0.203 \\
Gradient Boosting & 3 & 0.804 & 0.800 & 0.571 & 0.667 & 0.152 \\
Gradient Boosting & 4 & 0.833 & 0.818 & 0.346 & 0.486 & 0.130 \\
Gradient Boosting & 5 & 0.625 & 0.571 & 0.333 & 0.421 & 0.247 \\
Gradient Boosting & 7 & 0.761 & 0.895 & 0.315 & 0.466 & 0.171 \\

XGBoost & 1 & 0.768 & 0.792 & 0.362 & 0.497 & 0.161 \\
XGBoost & 2 & 0.730 & 0.755 & 0.544 & 0.632 & 0.207 \\
XGBoost & 3 & 0.759 & 0.400 & 0.571 & 0.471 & 0.196 \\
XGBoost & 4 & 0.755 & 0.714 & 0.385 & 0.500 & 0.144 \\
XGBoost & 5 & 0.639 & 0.625 & 0.417 & 0.500 & 0.259 \\
XGBoost & 7 & 0.772 & 0.720 & 0.333 & 0.456 & 0.171 \\

LightGBM & 1 & 0.774 & 0.823 & 0.357 & 0.498 & 0.159 \\
LightGBM & 2 & 0.712 & 0.766 & 0.529 & 0.626 & 0.208 \\
LightGBM & 3 & 0.777 & 0.500 & 0.714 & 0.588 & 0.184 \\
LightGBM & 4 & 0.742 & 0.714 & 0.385 & 0.500 & 0.139 \\
LightGBM & 5 & 0.708 & 0.778 & 0.583 & 0.667 & 0.247 \\
LightGBM & 7 & 0.762 & 0.783 & 0.333 & 0.468 & 0.171 \\

Deep Learning & 1 & 0.779 & 0.795 & 0.369 & 0.504 & 0.158 \\
Deep Learning & 2 & 0.734 & 0.760 & 0.559 & 0.644 & 0.204 \\
Deep Learning & 3 & 0.830 & 0.545 & 0.857 & 0.667 & 0.180 \\
Deep Learning & 4 & 0.780 & 0.750 & 0.346 & 0.474 & 0.136 \\
Deep Learning & 5 & 0.639 & 0.700 & 0.583 & 0.636 & 0.210 \\
Deep Learning & 7 & 0.758 & 0.731 & 0.352 & 0.475 & 0.172 \\

TabNet & 1 & 0.666 & 0.778 & 0.253 & 0.382 & 0.188 \\
TabNet & 2 & 0.606 & 0.793 & 0.338 & 0.474 & 0.247 \\
TabNet & 3 & 0.518 & 0.333 & 0.143 & 0.200 & 0.227 \\
TabNet & 4 & 0.652 & 0.889 & 0.308 & 0.457 & 0.157 \\
TabNet & 5 & 0.750 & 1.000 & 0.500 & 0.667 & 0.194 \\
TabNet & 7 & 0.571 & 0.579 & 0.204 & 0.301 & 0.222 \\
\hline
\end{tabular}
\end{minipage}
\hfill
\begin{minipage}{0.48\linewidth}
\centering
\caption{Fairness by poverty level (group-conditional performance).}
\label{tab:fairness_poverty}
\begin{tabular}{llcccccc}
\hline
Model & Group & AUC & Precision & Recall & F1 & Brier \\
\hline
Logistic Regression & 1 & 0.775 & 0.770 & 0.448 & 0.566 & 0.192 \\
Logistic Regression & 2 & 0.775 & 0.940 & 0.367 & 0.528 & 0.180 \\
Logistic Regression & 3 & 0.766 & 0.788 & 0.326 & 0.462 & 0.170 \\
Logistic Regression & 4 & 0.773 & 0.849 & 0.339 & 0.484 & 0.135 \\

Random Forest & 1 & 0.807 & 0.870 & 0.381 & 0.530 & 0.187 \\
Random Forest & 2 & 0.757 & 0.979 & 0.359 & 0.526 & 0.183 \\
Random Forest & 3 & 0.722 & 0.785 & 0.321 & 0.456 & 0.179 \\
Random Forest & 4 & 0.737 & 0.753 & 0.350 & 0.478 & 0.144 \\

Gradient Boosting & 1 & 0.766 & 0.705 & 0.410 & 0.518 & 0.196 \\
Gradient Boosting & 2 & 0.783 & 0.979 & 0.359 & 0.526 & 0.178 \\
Gradient Boosting & 3 & 0.749 & 0.795 & 0.321 & 0.458 & 0.172 \\
Gradient Boosting & 4 & 0.770 & 0.849 & 0.339 & 0.484 & 0.136 \\

XGBoost & 1 & 0.758 & 0.686 & 0.457 & 0.549 & 0.201 \\
XGBoost & 2 & 0.747 & 0.860 & 0.383 & 0.530 & 0.189 \\
XGBoost & 3 & 0.724 & 0.724 & 0.368 & 0.488 & 0.178 \\
XGBoost & 4 & 0.768 & 0.795 & 0.361 & 0.496 & 0.137 \\

LightGBM & 1 & 0.772 & 0.721 & 0.467 & 0.566 & 0.195 \\
LightGBM & 2 & 0.755 & 0.864 & 0.398 & 0.545 & 0.185 \\
LightGBM & 3 & 0.732 & 0.763 & 0.368 & 0.497 & 0.174 \\
LightGBM & 4 & 0.760 & 0.840 & 0.344 & 0.488 & 0.137 \\

Deep Learning & 1 & 0.779 & 0.707 & 0.552 & 0.620 & 0.188 \\
Deep Learning & 2 & 0.770 & 0.831 & 0.383 & 0.524 & 0.183 \\
Deep Learning & 3 & 0.750 & 0.761 & 0.363 & 0.491 & 0.172 \\
Deep Learning & 4 & 0.763 & 0.802 & 0.355 & 0.492 & 0.139 \\

TabNet & 1 & 0.638 & 0.784 & 0.276 & 0.408 & 0.241 \\
TabNet & 2 & 0.584 & 0.700 & 0.273 & 0.393 & 0.234 \\
TabNet & 3 & 0.648 & 0.764 & 0.285 & 0.415 & 0.200 \\
TabNet & 4 & 0.695 & 0.824 & 0.230 & 0.359 & 0.159 \\
\hline
\end{tabular}
\end{minipage}
\vspace{-8mm}
\end{table}

\section{Discussion}
\label{discussion}

In this study, we compare classical statistical models, tree-based machine-learning methods, and deep-learning architectures for predicting childhood overweight and obesity using nationally representative NSCH data. Across all approaches, performance converges within a narrow range, with AUC values clustering in the mid-to-high 0.70s. No model dominates across discrimination, calibration, sensitivity, and subgroup stability, indicating that predictive capacity is constrained more by the structure, resolution, and noise of survey-based epidemiologic data than by algorithmic complexity \cite{Christodoulou2019-od}.

The deep-learning model shows a distinct trade-off, achieving the highest recall and F1 score while maintaining discrimination comparable to boosting methods. Calibration degradation remains modest, suggesting improved sensitivity without severe loss of probability reliability\cite{Van_Calster2019-yw}. This makes the model potentially useful for screening settings where missing true cases is more harmful than generating false positives, but it is less suitable for precise individual risk prediction.

TabNet consistently underperforms across discrimination, recall, and calibration, indicating that sparse attention-based tabular architectures are poorly suited to noisy survey data.

Feature-importance analyses show strong agreement across models. Parental concern about child weight, physical activity, screen time, sleep, socioeconomic status, parental education, household stress, race, and neighborhood conditions repeatedly emerge as influential predictors across logistic regression, LightGBM, and TabNet. This convergence supports a multilevel structure of obesity risk spanning behavioral, household, and community domains, though importance measures remain descriptive rather than causal \cite{Sahoo2015-iu}.

Fairness evaluation reveals persistent disparities across race and poverty groups that remain largely invariant to model choice. Deep learning improves recall for some disadvantaged subgroups but does not eliminate disparities and sometimes increases calibration error, highlighting a central fairness trade-off between sensitivity and reliability. No model fully resolves subgroup inequities, highlighting that algorithmic advances alone are insufficient without improvements in data quality, representation, and measurement.

Overall, model choice should be guided by analytic objectives rather than marginal performance differences. Logistic regression and boosting methods provide stable, interpretable, and well-calibrated estimates for surveillance and policy analysis, while deep learning may be preferable when sensitivity is prioritized. Persistent fairness gaps across all models emphasize the need to address upstream data limitations and structural inequities rather than relying on increasing model complexity.

\section{Limitations}
\label{Limitation}

This study has several important limitations that constrain interpretation of the findings. 

We define the outcome using a binary classification of overweight or obese status and exclude underweight children. Although this definition aligns with public-health surveillance practices and simplifies model comparison, it reduces outcome granularity and likely contributes to the consistently modest recall observed across models. The limited sensitivity seen even in flexible algorithms suggests that meaningful variation in weight status may be compressed by categorical thresholds rather than constrained by model capacity.

Also, we prioritize model comparability and interpretability over aggressive optimization. We deliberately restrict hyperparameter tuning across models, which may disadvantage approaches such as deep learning and TabNet that typically require extensive tuning and larger effective sample sizes to perform optimally. In addition, we do not conduct ablation studies for neural network architectures, limiting insight into the contribution of individual model design components.

Moreover, our fairness evaluation remains descriptive rather than corrective. While we document systematic performance disparities across race, poverty level, sex, and metropolitan status, we do not apply algorithmic mitigation strategies such as reweighting, subgroup-specific decision thresholds, or fairness-constrained optimization. As a result, the analysis characterizes equity limitations in population-level obesity prediction but does not attempt to resolve them.

Furthermore, a data-related limitation concerns the variables \texttt{vegetables\_21}, \texttt{fruit\_21}, \texttt{SugarDrink\_21},
 which contains structurally coded filler values rather than meaningful variation in reported intake for the analytic sample. This limits its substantive interpretability and reduces confidence in dietary-related findings. 

\section{Conclusion}
\label{Conclusion}

In this study, we compare classical statistical models, tree-based machine-learning methods, and deep-learning architectures for predicting overweight and obesity among U.S. children aged 10–17 years using nationally representative survey data. Across all approaches, model performance is very similar, suggesting that prediction accuracy is limited mainly by how the survey data are structured and measured, rather than by how complex the modeling method is. Logistic regression and boosting methods provide strong discrimination and stable calibration, supporting their continued use in population-level surveillance and policy analysis, while deep learning modestly improves recall and F1 score without delivering decisive overall gains. TabNet underperforms across discrimination, calibration, and subgroup stability, suggesting limited suitability for noisy, correlated survey data. Feature-importance analyses consistently highlight predictors across behavioral, household, and community domains, including parental concern about child weight, physical activity, screen exposure, sleep, socioeconomic status, parental education, household stress, and neighborhood safety, reinforcing the multilevel nature of childhood obesity risk. Fairness evaluation reveals persistent disparities across race and poverty groups that remain largely invariant to model choice, indicating that equity limitations primarily reflect structural data characteristics rather than algorithmic bias, with sex-based and metro–rural differences comparatively small.

From a public-health perspective, these findings emphasize that model selection should be guided by analytic purpose rather than marginal performance gains. Logistic regression and boosting models provide transparent, well-calibrated estimates appropriate for surveillance, resource allocation, and policy analysis, while deep learning may be better suited for screening contexts where sensitivity is prioritized. However, improving predictive equity and sensitivity will require investments in data quality rather than increasingly complex models.

For researchers and policymakers, future efforts should focus on improvements in data quality and representation rather than increased model complexity. 
Using data that follow children over time, include objective health measurements, or link survey responses to administrative records may reduce measurement noise and improve predictive accuracy. Also, addressing inequities during data collection, such as improving representation and measurement quality, is likely to be more effective for improving both model performance and health equity than relying only on algorithmic adjustments after modeling.




\bibliography{Bibliography}

@ARTICLE{Kivrak2021-dl,
  title   = "Deep learning-based prediction of obesity levels according to
             eating habits and physical condition",
  author  = "Kivrak, Mehmet",
  journal = "The Journal of Cognitive Systems",
  volume  =  6,
  number  =  1,
  pages   = "24--27",
  year    =  2021
}

@MISC{Alvarez2011-ry,
  title        = "Prevalencia de sobrepeso y obesidad, y factores de riesgo, en
                  ni{\~n}os de 7-12 a{\~n}os, en una escuela p{\'u}blica de
                  Cartagena septiembre- octubre de 2010",
  booktitle    = "Edu.co",
  author       = "{\'A}lvarez, Hern{\'a}ndez and Mar{\'\i}a, Guiomar",
  year         =  2011,
  howpublished = "\url{https://repositorio.unal.edu.co/handle/unal/7739}",
  note         = "Accessed: 2025-12-15",
  language     = "es"
}

@MISC{obesity_overweight,
  title        = "Obesity and overweight",
  booktitle    = "Who.int",
  howpublished = "\url{https://www.who.int/news-room/fact-sheets/detail/obesity-and-overweight}",
  note         = "Accessed: 2025-12-15",
  language     = "en"
}

@BOOK{Oecd2019-ye,
  title     = "The heavy burden of obesity: The economics of prevention",
  author    = "{OECD}",
  publisher = "OECD Publishing",
  year      =  2019,
  language  = "en"
}

@ARTICLE{Colditz1999-aw,
  title    = "Economic costs of obesity and inactivity",
  author   = "Colditz, G A",
  journal  = "Med. Sci. Sports Exerc.",
  volume   =  31,
  number   = "11 Suppl",
  pages    = "S663--7",
  year     =  1999,
  language = "en"
}

@MISC{CDCChildObesity,
  title        = "Obesity and Overweight",
  booktitle    = "{CDC} National Center for Health Statistics",
  month        =  oct,
  year         =  2024,
  howpublished = "\url{https://www.cdc.gov/nchs/fastats/obesity-overweight.htm}",
  note         = "Accessed: 2025-12-15",
  language     = "en"
}

@MISC{CDCAdultObesity,
  title        = "Adult obesity facts",
  author       = "{CDC}",
  month        =  may,
  year         =  2024,
  howpublished = "\url{https://www.cdc.gov/obesity/adult-obesity-facts/index.html}",
  note         = "Accessed: 2025-12-15",
  language     = "en"
}

@MISC{nsch,
  title        = "About the National Survey of children's health",
  booktitle    = "Childhealthdata.org",
  howpublished = "\url{https://www.childhealthdata.org/learn-about-the-nsch/NSCH}",
  note         = "Accessed: 2025-12-15"
}

@article{Colmenarejo2020-ml,
  title={Machine Learning Models to Predict Childhood and Adolescent Obesity and Related Outcomes},
  author={Colmenarejo, G.},
  journal={PMC},
  year={2020},
  url={https://www.ncbi.nlm.nih.gov/pmc/articles/PMC7469049/}
}

@ARTICLE{Sun2024-ml,
  title    = "Using interpretable machine learning methods to identify the
              relative importance of lifestyle factors for overweight and
              obesity in adults: pooled evidence from {CHNS} and {NHANES}",
  author   = "Sun, Zhiyuan and Yuan, Yunhao and Farrahi, Vahid and Herold,
              Fabian and Xia, Zhengwang and Xiong, Xuan and Qiao, Zhiyuan and
              Shi, Yifan and Yang, Yahui and Qi, Kai and Liu, Yufei and Xu,
              Decheng and Zou, Liye and Chen, Aiguo",
  journal  = "BMC Public Health",
  volume   =  24,
  number   =  1,
  pages    = "3034",
  year     =  2024,
  language = "en"
}

@ARTICLE{Ganie2025-hy,
  title    = "Lifestyle data-based multiclass obesity prediction with
              interpretable ensemble models incorporating {SHAP} and {LIME}
              analysis",
  author   = "Ganie, Shahid Mohammad and Pramanik, Pijush Kanti Dutta and Zhao,
              Zhongming",
  journal  = "Sci. Rep.",
  volume   =  15,
  number   =  1,
  pages    = "36916",
  year     =  2025,
  language = "en"
}

@article{Gupta2019-dl,
   title={Obesity Prediction with EHR Data: A Deep Learning Approach with Interpretable Elements},
   volume={3},
   ISSN={2637-8051},
   url={http://dx.doi.org/10.1145/3506719},
   DOI={10.1145/3506719},
   number={3},
   journal={ACM Transactions on Computing for Healthcare},
   publisher={Association for Computing Machinery (ACM)},
   author={Gupta, Mehak and Phan, Thao-Ly T. and Bunnell, H. Timothy and Beheshti, Rahmatollah},
   year={2022},
   month=apr, pages={1–19} }

@ARTICLE{An2022-ai,
  title    = "Applications of artificial intelligence to obesity research:
              Scoping review of methodologies",
  author   = "An, Ruopeng and Shen, Jing and Xiao, Yunyu",
  journal  = "J. Med. Internet Res.",
  volume   =  24,
  number   =  12,
  pages    = "e40589",
  year     =  2022,
  language = "en"
}

@ARTICLE{Gulu2022-lu,
  title    = "Investigation of obesity, eating behaviors and physical activity
              levels living in rural and urban areas during the covid-19
              pandemic era: a study of Turkish adolescent",
  author   = "G{\"u}l{\"u}, Mehmet and Yapici, Hakan and Mainer-Pardos, Elena
              and Alves, Ana Ruivo and Nobari, Hadi",
  journal  = "BMC Pediatr.",
  volume   =  22,
  number   =  1,
  pages    = "405",
  year     =  2022,
  language = "en"
}

@ARTICLE{Reidpath2002-zu,
  title    = "An ecological study of the relationship between social and
              environmental determinants of obesity",
  author   = "Reidpath, Daniel D and Burns, Cate and Garrard, Jan and Mahoney,
              Mary and Townsend, Mardie",
  journal  = "Health Place",
  volume   =  8,
  number   =  2,
  pages    = "141--145",
  year     =  2002,
  language = "en"
}

@ARTICLE{Cohen2006-db,
  title    = "Collective efficacy and obesity: the potential influence of
              social factors on health",
  author   = "Cohen, Deborah A and Finch, Brian K and Bower, Aimee and Sastry,
              Narayan",
  journal  = "Soc. Sci. Med.",
  volume   =  62,
  number   =  3,
  pages    = "769--778",
  year     =  2006,
  language = "en"
}

@ARTICLE{Lamerz2005-bq,
  title    = "Social class, parental education, and obesity prevalence in a
              study of six-year-old children in Germany",
  author   = "Lamerz, A and Kuepper-Nybelen, J and Wehle, C and Bruning, N and
              Trost-Brinkhues, G and Brenner, H and Hebebrand, J and
              Herpertz-Dahlmann, B",
  journal  = "Int. J. Obes. (Lond)",
  volume   =  29,
  number   =  4,
  pages    = "373--380",
  year     =  2005,
  language = "en"
}

@ARTICLE{Ng2014-rs,
  title    = "Global, regional, and national prevalence of overweight and
              obesity in children and adults during 1980-2013: a systematic
              analysis for the Global Burden of Disease Study 2013",
  author   = "Ng, Marie and Fleming, Tom and Robinson, Margaret and Thomson,
              Blake and Graetz, Nicholas and Margono, Christopher and Mullany,
              Erin C and Biryukov, Stan and Abbafati, Cristiana and Abera,
              Semaw Ferede and Abraham, Jerry P and Abu-Rmeileh, Niveen M E and
              Achoki, Tom and AlBuhairan, Fadia S and Alemu, Zewdie A and
              Alfonso, Rafael and Ali, Mohammed K and Ali, Raghib and Guzman,
              Nelson Alvis and Ammar, Walid and Anwari, Palwasha and Banerjee,
              Amitava and Barquera, Simon and Basu, Sanjay and Bennett, Derrick
              A and Bhutta, Zulfiqar and Blore, Jed and Cabral, Norberto and
              Nonato, Ismael Campos and Chang, Jung-Chen and Chowdhury, Rajiv
              and Courville, Karen J and Criqui, Michael H and Cundiff, David K
              and Dabhadkar, Kaustubh C and Dandona, Lalit and Davis, Adrian
              and Dayama, Anand and Dharmaratne, Samath D and Ding, Eric L and
              Durrani, Adnan M and Esteghamati, Alireza and Farzadfar, Farshad
              and Fay, Derek F J and Feigin, Valery L and Flaxman, Abraham and
              Forouzanfar, Mohammad H and Goto, Atsushi and Green, Mark A and
              Gupta, Rajeev and Hafezi-Nejad, Nima and Hankey, Graeme J and
              Harewood, Heather C and Havmoeller, Rasmus and Hay, Simon and
              Hernandez, Lucia and Husseini, Abdullatif and Idrisov, Bulat T
              and Ikeda, Nayu and Islami, Farhad and Jahangir, Eiman and
              Jassal, Simerjot K and Jee, Sun Ha and Jeffreys, Mona and Jonas,
              Jost B and Kabagambe, Edmond K and Khalifa, Shams Eldin Ali
              Hassan and Kengne, Andre Pascal and Khader, Yousef Saleh and
              Khang, Young-Ho and Kim, Daniel and Kimokoti, Ruth W and Kinge,
              Jonas M and Kokubo, Yoshihiro and Kosen, Soewarta and Kwan, Gene
              and Lai, Taavi and Leinsalu, Mall and Li, Yichong and Liang,
              Xiaofeng and Liu, Shiwei and Logroscino, Giancarlo and Lotufo,
              Paulo A and Lu, Yuan and Ma, Jixiang and Mainoo, Nana Kwaku and
              Mensah, George A and Merriman, Tony R and Mokdad, Ali H and
              Moschandreas, Joanna and Naghavi, Mohsen and Naheed, Aliya and
              Nand, Devina and Narayan, K M Venkat and Nelson, Erica Leigh and
              Neuhouser, Marian L and Nisar, Muhammad Imran and Ohkubo,
              Takayoshi and Oti, Samuel O and Pedroza, Andrea and Prabhakaran,
              Dorairaj and Roy, Nobhojit and Sampson, Uchechukwu and Seo,
              Hyeyoung and Sepanlou, Sadaf G and Shibuya, Kenji and Shiri,
              Rahman and Shiue, Ivy and Singh, Gitanjali M and Singh, Jasvinder
              A and Skirbekk, Vegard and Stapelberg, Nicolas J C and Sturua,
              Lela and Sykes, Bryan L and Tobias, Martin and Tran, Bach X and
              Trasande, Leonardo and Toyoshima, Hideaki and van de Vijver,
              Steven and Vasankari, Tommi J and Veerman, J Lennert and
              Velasquez-Melendez, Gustavo and Vlassov, Vasiliy Victorovich and
              Vollset, Stein Emil and Vos, Theo and Wang, Claire and Wang,
              Xiaorong and Weiderpass, Elisabete and Werdecker, Andrea and
              Wright, Jonathan L and Yang, Y Claire and Yatsuya, Hiroshi and
              Yoon, Jihyun and Yoon, Seok-Jun and Zhao, Yong and Zhou, Maigeng
              and Zhu, Shankuan and Lopez, Alan D and Murray, Christopher J L
              and Gakidou, Emmanuela",
  journal  = "Lancet",
  volume   =  384,
  number   =  9945,
  pages    = "766--781",
  year     =  2014,
  language = "en"
}

@ARTICLE{Ford2008-ji,
  title    = "Epidemiology of obesity in the Western Hemisphere",
  author   = "Ford, Earl S and Mokdad, Ali H",
  journal  = "J. Clin. Endocrinol. Metab.",
  volume   =  93,
  number   = "11 Suppl 1",
  pages    = "S1--8",
  year     =  2008,
  language = "en"
}

@ARTICLE{Fontaine2003-ol,
  title    = "Years of life lost due to obesity",
  author   = "Fontaine, Kevin R and Redden, David T and Wang, Chenxi and
              Westfall, Andrew O and Allison, David B",
  journal  = "JAMA",
  volume   =  289,
  number   =  2,
  pages    = "187--193",
  year     =  2003,
  language = "en"
}

@ARTICLE{Berrington_de_Gonzalez2010-uf,
  title    = "Body-mass index and mortality among 1.46 million white adults",
  author   = "Berrington de Gonzalez, Amy and Hartge, Patricia and Cerhan,
              James R and Flint, Alan J and Hannan, Lindsay and MacInnis,
              Robert J and Moore, Steven C and Tobias, Geoffrey S and
              Anton-Culver, Hoda and Freeman, Laura Beane and Beeson, W
              Lawrence and Clipp, Sandra L and English, Dallas R and Folsom,
              Aaron R and Freedman, D Michal and Giles, Graham and Hakansson,
              Niclas and Henderson, Katherine D and Hoffman-Bolton, Judith and
              Hoppin, Jane A and Koenig, Karen L and Lee, I-Min and Linet,
              Martha S and Park, Yikyung and Pocobelli, Gaia and Schatzkin,
              Arthur and Sesso, Howard D and Weiderpass, Elisabete and Willcox,
              Bradley J and Wolk, Alicja and Zeleniuch-Jacquotte, Anne and
              Willett, Walter C and Thun, Michael J",
  journal  = "N. Engl. J. Med.",
  volume   =  363,
  number   =  23,
  pages    = "2211--2219",
  year     =  2010,
  language = "en"
}

@ARTICLE{Prospective_Studies_Collaboration2009-te,
  title    = "Body-mass index and cause-specific mortality in 900 000 adults:
              collaborative analyses of 57 prospective studies",
  author   = "{Prospective Studies Collaboration} and Whitlock, Gary and
              Lewington, Sarah and Sherliker, Paul and Clarke, Robert and
              Emberson, Jonathan and Halsey, Jim and Qizilbash, Nawab and
              Collins, Rory and Peto, Richard",
  journal  = "Lancet",
  volume   =  373,
  number   =  9669,
  pages    = "1083--1096",
  year     =  2009,
  language = "en"
}

@ARTICLE{Pischon2008-ol,
  title    = "General and abdominal adiposity and risk of death in Europe",
  author   = "Pischon, T and Boeing, H and Hoffmann, K and Bergmann, M and
              Schulze, M B and Overvad, K and van der Schouw, Y T and Spencer,
              E and Moons, K G M and Tj{\o}nneland, A and Halkjaer, J and
              Jensen, M K and Stegger, J and Clavel-Chapelon, F and
              Boutron-Ruault, M-C and Chajes, V and Linseisen, J and Kaaks, R
              and Trichopoulou, A and Trichopoulos, D and Bamia, C and Sieri, S
              and Palli, D and Tumino, R and Vineis, P and Panico, S and
              Peeters, P H M and May, A M and Bueno-de-Mesquita, H B and van
              Duijnhoven, F J B and Hallmans, G and Weinehall, L and Manjer, J
              and Hedblad, B and Lund, E and Agudo, A and Arriola, L and
              Barricarte, A and Navarro, C and Martinez, C and Quir{\'o}s, J R
              and Key, T and Bingham, S and Khaw, K T and Boffetta, P and
              Jenab, M and Ferrari, P and Riboli, E",
  journal  = "N. Engl. J. Med.",
  volume   =  359,
  number   =  20,
  pages    = "2105--2120",
  year     =  2008,
  language = "en"
}

@ARTICLE{Jaacks2019-ur,
  title    = "The obesity transition: stages of the global epidemic",
  author   = "Jaacks, Lindsay M and Vandevijvere, Stefanie and Pan, An and
              McGowan, Craig J and Wallace, Chelsea and Imamura, Fumiaki and
              Mozaffarian, Dariush and Swinburn, Boyd and Ezzati, Majid",
  journal  = "Lancet Diabetes Endocrinol.",
  volume   =  7,
  number   =  3,
  pages    = "231--240",
  year     =  2019,
  language = "en"
}

@ARTICLE{Lavie2014-ah,
  title   = "Obesity and cardiovascular diseases: Implications regarding
             fitness, fatness, and severity in the obesity paradox",
  author  = "Lavie, C J and Mcauley, P A and Church, T S and Milani, R V and
             Blair, S N",
  journal = "J. Am. Coll. Cardiol",
  volume  =  63,
  pages   = "1345--1354",
  year    =  2014
}

@ARTICLE{Koh-Banerjee2004-zf,
  title    = "Changes in whole-grain, bran, and cereal fiber consumption in
              relation to 8-y weight gain among men",
  author   = "Koh-Banerjee, Pauline and Franz, Mary and Sampson, Laura and Liu,
              Simin and Jacobs, Jr, David R and Spiegelman, Donna and Willett,
              Walter and Rimm, Eric",
  journal  = "Am. J. Clin. Nutr.",
  volume   =  80,
  number   =  5,
  pages    = "1237--1245",
  year     =  2004,
  language = "en"
}

@ARTICLE{Ford2005-ml,
  title    = "The epidemiology of obesity and asthma",
  author   = "Ford, Earl S",
  journal  = "J. Allergy Clin. Immunol.",
  volume   =  115,
  number   =  5,
  pages    = "897--909; quiz 910",
  year     =  2005,
  language = "en"
}

@ARTICLE{Bakhshi2015-nu,
  title    = "Obesity and related factors in Iran: The {STEPS} Survey, 2011",
  author   = "Bakhshi, Enayatollah and Koohpayehzadeh, Jalil and Seifi, Behjat
              and Rafei, Ali and Biglarian, Akbar and Asgari, Fereshteh and
              Etemad, Koorosh and Bidhendi Yarandi, Razieh",
  journal  = "Iran. Red Crescent Med. J.",
  volume   =  17,
  number   =  6,
  pages    = "e22479",
  year     =  2015,
  language = "en"
}

@ARTICLE{Sarlio-Lahteenkorva2004-bi,
  title    = "Relative weight and income at different levels of socioeconomic
              status",
  author   = "Sarlio-L{\"a}hteenkorva, Sirpa and Silventoinen, Karri and
              Lahelma, Eero",
  journal  = "Am. J. Public Health",
  volume   =  94,
  number   =  3,
  pages    = "468--472",
  year     =  2004,
  language = "en"
}

@ARTICLE{Bonauto2014-nb,
  title   = "Peer reviewed: Obesity prevalence by occupation in Washington
             State, behavioral risk factor surveillance system",
  author  = "Bonauto, D K and Lu, D and Fan, Z J",
  journal = "Prev. Chronic Dis",
  year    =  2014
}

@ARTICLE{John2005-mo,
  title    = "Smoking status, cigarettes per day, and their relationship to
              overweight and obesity among former and current smokers in a
              national adult general population sample",
  author   = "John, U and Hanke, M and Rumpf, H-J and Thyrian, J R",
  journal  = "Int. J. Obes. (Lond)",
  volume   =  29,
  number   =  10,
  pages    = "1289--1294",
  year     =  2005,
  language = "en"
}

@ARTICLE{Besson2009-wk,
  title    = "A cross-sectional analysis of physical activity and obesity
              indicators in European participants of the {EPIC-PANACEA} study",
  author   = "Besson, H and Ekelund, U and Luan, J and May, A M and Sharp, S
              and Travier, N and Agudo, A and Slimani, N and Rinaldi, S and
              Jenab, M and Norat, T and Mouw, T and Rohrmann, S and Kaaks, R
              and Bergmann, M and Boeing, H and Clavel-Chapelon, F and
              Boutron-Ruault, M C and Overvad, K and Andreasen, E L and
              Johnsen, N F{\o}ns and Halkjaer, J and Gonzalez, C and Rodriguez,
              L and Sanchez, M J and Arriola, L and Barricarte, A and Navarro,
              C and Key, T J and Spencer, E A and Orfanos, P and Naska, A and
              Trichopoulou, A and Manjer, J and Wirf{\"a}lt, E and Lund, E and
              Palli, D and Agnoli, C and Vineis, P and Panico, S and Tumino, R
              and Bueno-de-Mesquita, H B and van den Berg, S W and Odysseos, A
              D and Riboli, E and Wareham, N J and Peeters, P H",
  journal  = "Int. J. Obes. (Lond)",
  volume   =  33,
  number   =  4,
  pages    = "497--506",
  year     =  2009,
  language = "en"
}

@ARTICLE{Popkin2012-br,
  title    = "Global nutrition transition and the pandemic of obesity in
              developing countries",
  author   = "Popkin, Barry M and Adair, Linda S and Ng, Shu Wen",
  journal  = "Nutr. Rev.",
  volume   =  70,
  number   =  1,
  pages    = "3--21",
  year     =  2012,
  language = "en"
}

@ARTICLE{Kahan2016-qd,
  title    = "Overweight and obesity management strategies",
  author   = "Kahan, Scott",
  journal  = "Am. J. Manag. Care",
  volume   =  22,
  number   = "7 Suppl",
  pages    = "s186--96",
  year     =  2016,
  language = "en"
}

@ARTICLE{Van_Baal2008-oq,
  title    = "Lifetime medical costs of obesity: prevention no cure for
              increasing health expenditure",
  author   = "van Baal, Pieter H M and Polder, Johan J and de Wit, G Ardine and
              Hoogenveen, Rudolf T and Feenstra, Talitha L and Boshuizen,
              Hendriek C and Engelfriet, Peter M and Brouwer, Werner B F",
  journal  = "PLoS Med.",
  volume   =  5,
  number   =  2,
  pages    = "e29",
  year     =  2008,
  language = "en"
}

@ARTICLE{Spiegelman2001-ij,
  title    = "Obesity and the regulation of energy balance",
  author   = "Spiegelman, B M and Flier, J S",
  journal  = "Cell",
  volume   =  104,
  number   =  4,
  pages    = "531--543",
  year     =  2001,
  language = "en"
}

@misc{tabnet,
      title={TabNet: Attentive Interpretable Tabular Learning}, 
      author={Sercan O. Arik and Tomas Pfister},
      year={2020},
      eprint={1908.07442},
      archivePrefix={arXiv},
      primaryClass={cs.LG},
      url={https://arxiv.org/abs/1908.07442}, 
}

@MISC{Molnar2025-ja,
  title        = "Interpretable machine learning",
  author       = "Molnar, Christoph",
  month        =  mar,
  year         =  2025,
  howpublished = "\url{https://christophm.github.io/interpretable-ml-book/}",
  note         = "Accessed: 2025-12-15",
  language     = "en"
}

@ARTICLE{Christodoulou2019-od,
  title    = "A systematic review shows no performance benefit of machine
              learning over logistic regression for clinical prediction models",
  author   = "Christodoulou, Evangelia and Ma, Jie and Collins, Gary S and
              Steyerberg, Ewout W and Verbakel, Jan Y and Van Calster, Ben",
  journal  = "J. Clin. Epidemiol.",
  volume   =  110,
  pages    = "12--22",
  year     =  2019,
  language = "en"
}

@ARTICLE{Van_Calster2019-yw,
  title    = "Calibration: the Achilles heel of predictive analytics",
  author   = "Van Calster, Ben and McLernon, David J and van Smeden, Maarten
              and Wynants, Laure and Steyerberg, Ewout W and {Topic Group
              `Evaluating diagnostic tests and prediction models' of the
              STRATOS initiative}",
  journal  = "BMC Med.",
  volume   =  17,
  number   =  1,
  pages    = "230",
  year     =  2019,
  language = "en"
}

@ARTICLE{Sahoo2015-iu,
  title    = "Childhood obesity: causes and consequences",
  author   = "Sahoo, Krushnapriya and Sahoo, Bishnupriya and Choudhury, Ashok
              Kumar and Sofi, Nighat Yasin and Kumar, Raman and Bhadoria, Ajeet
              Singh",
  journal  = "J. Family Med. Prim. Care",
  volume   =  4,
  number   =  2,
  pages    = "187--192",
  year     =  2015,
  language = "en"
}

\end{document}